\PassOptionsToPackage{capitalize}{cleveref}
\documentclass[10pt,twocolumn,letterpaper]{article}

\usepackage[pagenumbers]{wacv}
\usepackage{rotating,booktabs,multirow}

\definecolor{wacvblue}{rgb}{0.21,0.49,0.74}
\usepackage[pagebackref,breaklinks,colorlinks,allcolors=wacvblue]{hyperref}
\usepackage{cleveref}
\usepackage[table]{xcolor}

\definecolor{pearFour}{HTML}{588F27}
\definecolor{pearFith}{HTML}{ECF0F1}
\definecolor{pearDark}{HTML}{2980B9}
\definecolor{pearDarker}{HTML}{1D2DEC}
\definecolor{pearLight}{HTML}{F7E0D5}

\definecolor{eva01purple}{RGB}{168,119,200}
\definecolor{eva02red}{RGB}{236,35,35}

\usepackage{url}
\usepackage{nicefrac}
\usepackage{lipsum}
\usepackage{fancyhdr}
\usepackage{appendix}
\usepackage{afterpage}

\usepackage{graphicx}
\usepackage{caption}
\usepackage{wrapfig}
\usepackage{float}
\usepackage{soul}

\usepackage{booktabs}
\usepackage{multirow}
\usepackage{threeparttable}
\usepackage{arydshln}

\usepackage{amsmath}
\usepackage{amssymb}
\usepackage{amsfonts}
\usepackage{bm}
\usepackage{dsfont}

\usepackage{algorithm}
\usepackage{algorithmic}
\usepackage{listings}
\usepackage{verbatim}
\usepackage{comment}

\usepackage{etoolbox}
\makeatletter
\AfterEndEnvironment{algorithm}{\let\@algcomment\relax}
\AtEndEnvironment{algorithm}{\kern2pt\hrule\relax\vskip3pt\@algcomment}
\let\@algcomment\relax
\newcommand\algcomment[1]{\def\@algcomment{\footnotesize#1}}
\renewcommand\fs@ruled{\def\@fs@cfont{\bfseries}\let\@fs@capt\floatc@ruled
  \def\@fs@pre{\hrule height.8pt depth0pt \kern2pt}%
  \def\@fs@post{}%
  \def\@fs@mid{\kern2pt\hrule\kern2pt}%
  \let\@fs@iftopcapt\iftrue}
\makeatother

\usepackage{pifont}

\usepackage{textcomp}
\usepackage{relsize}
\usepackage{xspace}

\usepackage{tikz}
\usetikzlibrary{decorations.pathreplacing, positioning}

\usepackage{amsthm}
\usepackage{tcolorbox}
\tcbuselibrary{theorems,skins,breakable}

\newtcbtheorem
  [crefname={observation}{observations}]
  {observation}
  {Observation}
  {
    fontupper=\small,
    colback=wacvblue!7,
    colframe=wacvblue,
    colbacktitle=wacvblue,
    coltitle=white,
    fonttitle=\bfseries,
    boxsep=1pt,
    left=1.5mm,
    right=1.5mm,
    top=2mm,
    bottom=1mm,
  }
  {obs}

\newtcolorbox{takeaway}{
  enhanced, breakable,
  borderline west={2.5pt}{0pt}{wacvblue},
  colback=gray!8,
  colframe=white,
  fontupper=\small,
  left=6pt, right=4pt,
  top=3pt, bottom=3pt,
  before upper={%
    {\bfseries\color{wacvblue}\small Takeaway.}\enspace
  }
}
\usepackage{enumitem}

\newcommand{\ours}{FOX\xspace}

\def\wacvPaperID{****}
\def\confName{WACV}
\def\confYear{2027}

\makeatletter
\apptocmd\@maketitle{{\teaser{}}}{}{}
\makeatother
\newcommand{\teaser}{%
\vspace{-10pt}
\centering
\small
\includegraphics[width=\linewidth]{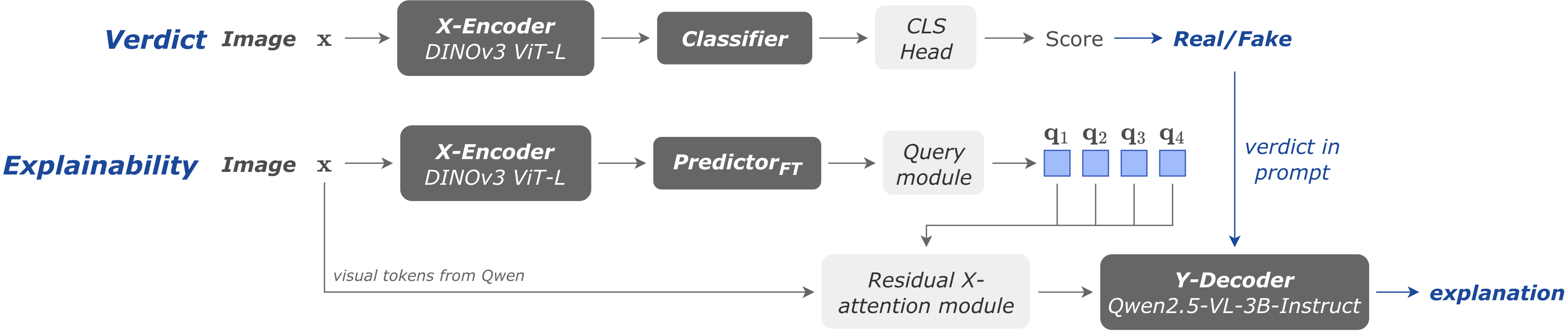}
\captionof{figure}{
\textbf{Inference in \ours.} \textbf{(Top)} Detection uses a pure-vision encoder: the verdict is read from the \texttt{Classifier} logits over the adapted \texttt{X-Encoder}, without language-model decoding. \textbf{(Bottom)} Explanation is separate. \texttt{Predictor}$_{\texttt{FT}}$ converts the detector's visual representation into caption-space forensic guidance that \texttt{Y-Decoder} verbalizes, conditioned on the locked verdict.}
\label{fig:teaser}
\par\vspace{10pt}
}

\title{
Generalizable AI-Generated Image Detection with \\Embedding-Space Explanations
}

\author{
Nikos Giakoumoglou$^{1}$\thanks{Corresponding author: \texttt{nikos@imperial.ac.uk}}
\quad
Paschalis Giakoumoglou$^{2}$
\quad
Tania Stathaki$^{1}$
\quad
Symeon Papadopoulos$^{2}$\\[0.5em]
$^{1}$Imperial College London \qquad
$^{2}$Information Technologies Institute, CERTH\vspace{.5em}\\
{Code: \url{https://github.com/mever-team/fox}}\\
}

\begin{document}
\maketitle

\begin{abstract}
AI-generated images have reached unprecedented realism, while instruction-driven AI editors re-synthesize images for local changes, leaving no splice boundary to localize. Most detectors are black boxes with unverifiable verdicts and generalize poorly because adapting a backbone on the detection distribution memorizes generator-specific fingerprints that quickly become obsolete. Specialist detectors built on pure-vision representations, particularly DINOv3, generalize well across generators, but explainable methods forgo this advantage: jointly training one multimodal model for classification and language generation subjects its visual representation to the language objective. We therefore require that explanation training cannot alter the verdict. We introduce \ours, which first builds and locks a specialist DINOv3 detector, then uses its representation to guide a lightweight VLM toward forensic explanation. A joint-embedding predictor maps locked visual evidence into a frozen text encoder's semantic space, producing guidance that the VLM verbalizes on demand. \ours achieves 97.4 mean AUC across 14 unseen generators, 69.8 BERTScore on XPlainVerse, and 56.1 on LOKI.
\end{abstract}
\section{Introduction}
\label{sec:introduction}

The evolution of GANs~\cite{goodfellow2014generative, karras2019style} and diffusion models~\cite{ho2020denoising, dhariwal2021diffusion, rombach2022high} has made synthetic images highly realistic, enabling misinformation, targeted attacks, and online harms through weaponised deepfake content. Recent diffusion models~\cite{flux2024,esser2024scaling,giakoumoglou2026harmful} and autoregressive techniques~\cite{tian2024visual} expand the threat surface: instruction-following editors built on VLMs perform targeted manipulations by re-synthesizing the entire image rather than splicing a region into it~\cite{gupta2025multifakeverse}. With no pasted patch, boundary inconsistency, or provenance mask, forensic localization operates near chance, and humans fare little better~\cite{gupta2025multifakeverse, giakoumoglou2025sagi}. A bare real-versus-fake score is then of limited use, since journalists, moderators, and users need to know \emph{why} an image was flagged.

Existing detectors exploit low-level generator traces~\cite{durall2020watch, frank2020leveraging, corvi2023detection, bammey2023synthbuster, tan2024npr, karageorgiou2025spai, giakoumoglou2026diffusionprint} or semantic representations~\cite{ojha2023towards, cozzolino2024raising, koutlis2024leveraging, sha2023fake, liu2024forgery}. Four issues remain. \textbf{First}, most detectors are black boxes, whose verdicts humans cannot verify. \textbf{Second}, unconstrained backbone adaptation memorizes generator-specific fingerprints and transfers poorly~\cite{zhu2023genimage, yan2024sanity, gragnaniello2021gan, chai2020makes, ojha2023towards}. \textbf{Third}, artifact-driven methods depend on high-frequency traces that improving generators mimic and resizing or compression destroy~\cite{corvi2023detection, karageorgiou2025spai}. \textbf{Fourth}, explainable MLLM detectors~\cite{xu2024fakeshield, huang2024sida, li2024forgerygpt, sun2024forgerysleuth, chen2025xdfd, huang2024ffaa} fine-tune billions of parameters for a vision-first problem, with weaker detection~\cite{jia2024can, ye2024loki, zhou2025aigiholmes}.

At the same time, a new line of forensic detectors has emerged around pure vision self-supervised encoders, particularly DINOv3~\cite{simeoni2025dinov3}. Frozen DINO representations generalize strongly across generators~\cite{huang2025rethinking}, parameter-efficient DINOv3 adaptations perform strongly in recent competitions~\cite{qu2026dinomac}, and DINOv3 systems dominate recent robustness evaluations~\cite{hopf2026robustdeepfakedetectionntire}. This suggests that reliable forensic perception may be better provided by a specialist pure vision encoder than by a general-purpose MLLM vision tower (\cref{subsec:related-dinov3}). Yet, explainable detection largely follows the opposite design: authenticity prediction and explanation generation are optimized together in one multimodal pipeline (\cref{subsec:related-mllm}), coupling objectives with different representations, losses, and convergence timescales.

We therefore ask: \emph{rather than making one model responsible for both perception and language, can we first obtain the strongest possible detector and use its representation to help a lightweight VLM explain the fixed verdict?} Explanation need not require the language model to rediscover forensic perception from scratch. A VLM already provides strong language generation and image understanding, but it lacks access to the specialist detector's forensic evidence. Decoupling the objectives lets detection optimize solely for cross-generator transfer and explanation optimize afterwards without altering the verdict.

\emph{Joint Embedding Predictive Architectures} (JEPAs)~\cite{lecun2022path} provide a natural interface between these modules by predicting target representations in latent space~\cite{assran2023ijepa, bardes2024vjepa, assran2025vjepa2, huang2026llmjepa, chen2026vljepa}. For explanation, this means predicting \emph{what the forensic evidence describes} instead of reconstructing an exact token realization. \ours predicts evidence in the semantic space of a frozen text encoder, where semantically equivalent explanations remain nearby despite different wording. The fixed target space removes \textit{target co-collapse}, where predictor and target encoder jointly converge to a constant~\cite{jing2022understanding, kenneweg2025jepa}, while contrastive learning discourages a constant predictor. A VLM then verbalizes the representation only when requested~\cite{alayrac2022flamingo, chen2026vljepa}; these embeddings provide forensic guidance without reconstructing the classifier's internal reasoning.

We instantiate this design in \ours (\textbf{\underline{FO}}rensic e\textbf{\underline{X}}planations) (\cref{fig:teaser}). We first investigate which visual representation preserves forensic evidence and how it should be adapted. Under a common readout, view-consistency self-supervised encoders transfer best, with DINOv3 being the strongest; brief LoRA adaptation selected on an external generator distribution further improves transfer (\cref{subsec:ablation-backbone,subsec:ablation-adaptation}). The detector is permanently locked. A separate predictor maps its visual features into a frozen text-embedding space, and predicted query vectors guide a Qwen2.5-VL-3B decoder. In that way, the VLM never determines authenticity: it only verbalizes the explanation  solely on the basis of the input image, fixed verdict, and detector-derived evidence. This separation lets detector and explainer use different objectives and optimization timescales while allowing language generation to benefit from the specialist representation.

Our main \textbf{contributions} are the following:
\begin{itemize}[nosep, leftmargin=*]
\item We introduce \ours, an explainable AI-generated image detector that decouples specialist visual detection from language generation: the verdict is completed and locked before any explanation objective, so explanation training cannot alter detection.
\item We systematically study the detection pathway from pretrained representations. Under an identical readout, DINOv3 transfers better than language-supervised, reconstruction-based, and latent-prediction alternatives, while external-validation checkpoint selection outperforms freezing, end-of-epoch LoRA, and internal-validation selection (\cref{subsec:ablation-backbone,subsec:ablation-adaptation}).
\item We show how the specialist representation guides a lightweight VLM. A joint-embedding predictor maps detector features into a frozen text space, and predicted query vectors condition the decoder toward forensic evidence without requiring the VLM to perform detection.
\item \ours reaches 97.4 mean AUC across 14 unseen generators (\cref{tab:detection}), 69.8 BERTScore on XPlainVerse (XPV) and 56.1 on LOKI (\cref{tab:explanation}), while remaining robust to common perturbations (\cref{fig:robustness}).
\end{itemize}
\vspace{-6pt}
\section{Related Work}
\label{sec:related-work}

\vspace{-4pt}
\subsection{AI-Generated Image Detection}
\label{subsec:related-aigi}

Low-level detectors exploit frequency traces~\cite{durall2020watch, frank2020leveraging, zhang2019detecting, corvi2023detection, bammey2023synthbuster, li2024masksim, doloriel2024frequency, tan2024frequency, giakoumoglou2026diffusionprint}, gradients~\cite{tan2023learning}, pixel relationships~\cite{tan2024npr}, or texture-rich patches~\cite{zhong2024patchcraft}; SPAI~\cite{karageorgiou2025spai} treats synthetic images as spectral outliers. Semantic methods probe or adapt vision-language features~\cite{ojha2023towards, koutlis2024leveraging, cozzolino2024raising, liu2024forgery, yan2024sanity, sha2023fake}, while reconstruction-based approaches score generative fidelity~\cite{wang2023dire, ricker2024aeroblade, cazenavette2024fakeinversion, luo2024lare, he2024rigid, cozzolino2025zero, chendrct}. Despite strong in-domain results, cross-generator transfer remains difficult~\cite{gragnaniello2021gan, chai2020makes, zhu2023genimage, mareen2026tgif2}, and most return only a score. 

\vspace{-5pt}
\subsection{Explainable Detection with MLLMs}
\label{subsec:related-mllm}

General-purpose MLLMs trail specialist detectors on bare classification~\cite{jia2024can, li2024fakebench, ye2024loki}, motivating supervised forensic variants spanning face forgery~\cite{zhang2024common, huang2024ffaa}, expert-feature fusion~\cite{chen2025xdfd}, segmentation/trace encoders~\cite{xu2024fakeshield, huang2024sida, li2024forgerygpt, sun2024forgerysleuth, kang2025legionlearninggroundexplain}, preference optimization~\cite{zhou2025aigiholmes}, and VQA-style formulations~\cite{chang2023antifakeprompt, keita2024bi}. Most train billions of parameters for a perception problem, many adapt the vision tower and so risk cross-generator transfer, and they generate explanations autoregressively together with the verdict, thereby coupling detection and explanation quality within one set of weights. \ours instead locks perception first, predicts forensic content in a fixed semantic space, and uses the VLM only to verbalize an already-decided verdict, thus avoiding the use of the language model as a forensic detector.

\subsection{Pure Vision Backbones for Forensics}
\label{subsec:related-dinov3}

The standard encoder was long a language-aligned model~\cite{radford2021learning, ojha2023towards, cozzolino2024raising, liu2023llava, wang2024qwen2}, but results now favour purely visual self-supervised encoders, specifically  DINOv2~\cite{konstantinidou2025itw} and DINOv3~\cite{simeoni2025dinov3}. Frozen probes generalize across generators~\cite{huang2025rethinking}, parameter-efficient adaptations win open challenges~\cite{qu2026dinomac}, and the latest robustness benchmark is led almost uniformly by DINOv3 entries~\cite{hopf2026robustdeepfakedetectionntire}. This says nothing about how much to adapt, since the frozen-probe~\cite{huang2025rethinking, ojha2023towards} and parameter-efficient~\cite{qu2026dinomac, liu2024forgery} lines reach opposite conclusions, and checkpoint selection remains vague or oracle-based. \Cref{sec:ablations} systematically examines both choices.
\section{Methodology}
\label{sec:method}
 
\begin{figure*}[!t]
\centering
\small
\includegraphics[width=0.98\linewidth]{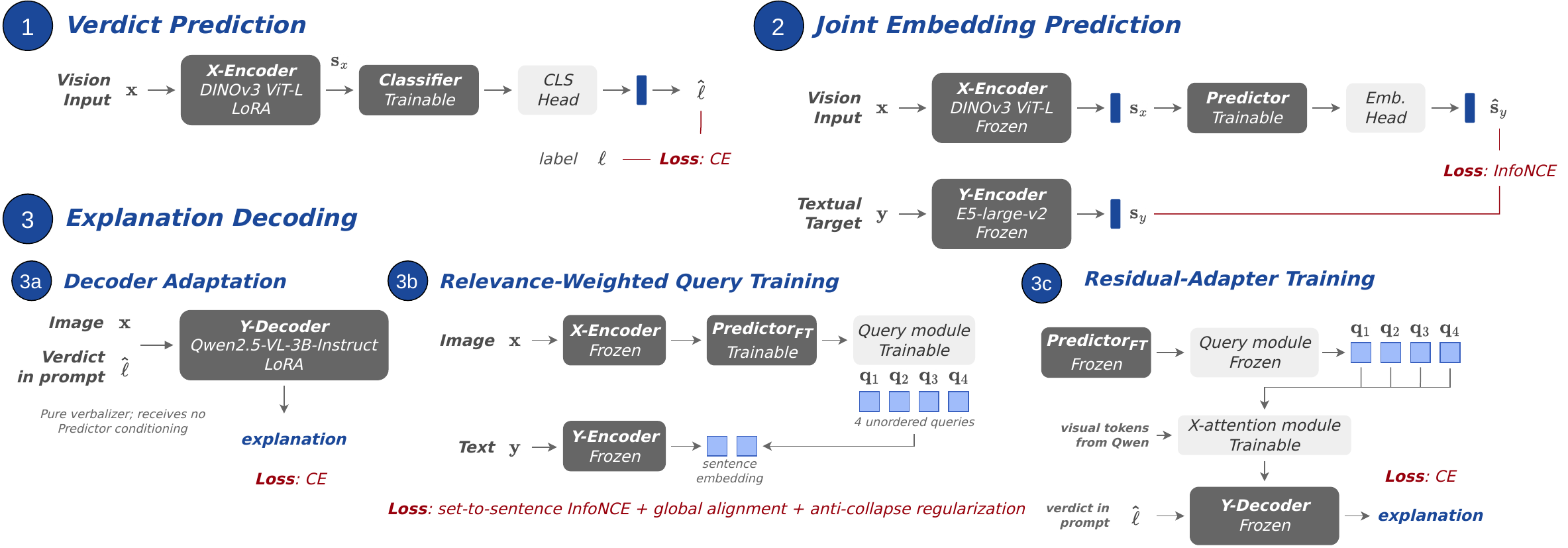}
\caption{\textbf{Training of \ours.} Detection and explanation are optimized separately. Stage~1 adapts the \texttt{X-Encoder} through LoRA jointly with the \texttt{Classifier} under cross-entropy; Stage~2 trains the \texttt{Predictor} with InfoNCE against the frozen \texttt{Y-Encoder}. Stage~3 trains a LoRA on the decoder (3a), the query module together with the \texttt{Predictor} (3b), and the residual adapter (3c).}
\label{fig:training}
\end{figure*}

We propose \ours, a \emph{Joint Embedding Predictive Architecture} for generalizable and explainable AI-generated image detection, realized as brief visual adaptation, a locked detector, forensic prediction in a fixed semantic embedding space, and on-demand verbalization. \ours is trained with triplets $\langle \mathbf{x}, \mathbf{y}, \ell \rangle$, where $\mathbf{x}$ is the visual input (a single image), $\mathbf{y}$ the textual target (the natural language explanation of its forensic evidence), and $\ell \in \{0, 1\}$ the authenticity label. Given a test image, the model must output a verdict $\hat{\ell}$ and, on demand, a human-readable explanation $\hat{\mathbf{y}}$ that presents forensic evidence supporting that verdict. Throughout, \emph{frozen} means a module receives no gradient in any stage, and \emph{locked} means it was trained in an earlier stage and receives none afterwards.
 
\ours comprises \textbf{five} components:
\begin{enumerate}
    \item \texttt{X-Encoder} $f_x$ ($\mathbf{x} \rightarrow \mathbf{s}_x$) compresses the input image into compact visual features. It is initialized from a pure vision self-supervised backbone with frozen weights, 
     adapted through LoRA in Stage 1, and locked thereafter.
    \item \texttt{Classifier} $c_\phi$ ($\mathbf{s}_x \rightarrow \hat{\ell}$) maps the visual features to the verdict $\hat{\ell}$ through a classification readout. It is trained jointly with the \texttt{X-Encoder} LoRA parameters and  locked at the end of Stage 1, so the verdict function is fixed before any explanation objective is initialized.
    \item \texttt{Predictor} $g_\theta$ ($\mathbf{s}_x \rightarrow \langle \hat{\mathbf{s}}_y, \mathbf{Q} \rangle$) maps the same locked visual features into the fixed semantic embedding space, producing a global prediction $\hat{\mathbf{s}}_y$ and, once extended with a query module, a set $\mathbf{Q}$ of caption-space vectors. It is optimized only after \texttt{Classifier} is locked, so its objective cannot alter the learned verdict.
    \item \texttt{Y-Encoder} $f_y$ ($\mathbf{y} \rightarrow \mathbf{s}_y$) embeds the textual target into the continuous space that serves as the prediction target, abstracting away surface information such as phrasing and word order. It is frozen, used only to precompute training targets, and discarded at inference.
    \item \texttt{Y-Decoder} $d_\psi$ ($\langle \mathbf{x}, \hat{\ell}, \mathbf{Q} \rangle \rightarrow \hat{\mathbf{y}}$) verbalizes the verdict into grounded natural language forensic evidence.
\end{enumerate}
\Cref{fig:training} illustrates how we instantiate \ours. The \texttt{X-Encoder} is DINOv3 ViT-L/16~\cite{simeoni2025dinov3}, adapted through a rank-32 LoRA; its block-wise readouts are projected and fed to \texttt{Classifier}, a lightweight transformer trunk~\cite{vaswani2017attention} with a classification readout. An independently parameterized \texttt{Predictor} of the same shape reads the same locked visual features and is trained against targets from \texttt{Y-Encoder}, a frozen E5-large-v2~\cite{wang2022e5}. \texttt{Y-Decoder} is Qwen2.5-VL-3B-Instruct~\cite{bai2025qwen25vl}, a late-fusion VLM whose vision encoder remains frozen and whose language side carries a lightweight LoRA adapter. Full component definitions are given in Appendix~\ref{appendix:components}.
 
\subsection{Training Objectives}
\label{subsec:method-objectives}
 
Training \ours proceeds in three sequential stages (\cref{fig:training}), each optimizing a disjoint subset of parameters while every previously trained component remains locked.
 

\noindent\textbf{Stage 1 (Verdict Prediction).}
Stage~1 jointly trains the \texttt{X-Encoder} LoRA and \texttt{Classifier} on image-label pairs $\langle \mathbf{x}, \ell \rangle$ (\cref{fig:teaser}), using \textbf{cross-entropy} on authenticity labels alone, without captions, text encoders, or explanation losses. Checkpoints are selected by AUC on Chameleon, an independently collected external validation set disjoint from XPV training. Its provenance and generalization to newer generators are further examined in Appendix~\ref{appendix:cham-generalization}. The selected encoder and classifier are then locked, fixing the verdict function for all subsequent stages of \ours.

\noindent\textbf{Stage 2 (Joint Embedding Prediction).} Stage 2 trains the \texttt{Predictor} on triplets $\langle \mathbf{x}, \mathbf{y}, \ell \rangle$ with the \texttt{X-Encoder} locked and the \texttt{Y-Encoder} frozen (\cref{fig:training}). The embedding readout is trained with \textbf{InfoNCE}~\cite{oord2018cpc} at temperature $\tau$, aligning each predicted embedding $\hat{\mathbf{s}}_y$ with the target embedding $\mathbf{s}_y$ of its own explanation while contrasting it against the other explanations in the batch. Because the \texttt{Y-Encoder} is frozen, the targets are fixed points in a pretrained space and cannot drift toward the predictor, so the trivial solution in which both sides converge to a constant is unavailable; the contrastive term additionally penalizes a constant predictor directly. It is also the better fit geometrically: pretrained text embeddings are strongly anisotropic, concentrating in a narrow cone in which unrelated pairs still exhibit high cosine similarity~\cite{ethayarajh2019contextual, gao2019representation}, so absolute similarity targets carry little signal whereas relative comparisons against in-batch negatives remain discriminative, and a contrastive objective is suitable for this geometry~\cite{gao2021simcse}. The resulting JEPA predictor therefore learns explanation-aligned forensic representations without changing the detector's verdict function.
 
\noindent\textbf{Stage 3 (Explanation Decoding).} Stage 3 trains the explanation pathway in three steps. First (Stage 3a), a language-side LoRA on \texttt{Y-Decoder} is trained with \textbf{cross-entropy} on the reference explanation tokens, conditioned only on the image and the locked verdict written into the prompt as a word (\eg, \textit{``This image is fake. Write the forensic explanation.''}); only these LoRA parameters are trained, and they remain locked for the rest of training and at inference. We refer to the resulting decoder, which receives no query conditioning, as the \emph{unconditioned decoder}. Second (Stage 3b), we attach to the trained \texttt{Predictor} a small cross-attention module with four learned query slots (\cref{subsec:ablation-queries}) that attend over its hidden state and block tokens to produce four query vectors $\mathbf{q}_1, \ldots, \mathbf{q}_4$ (\cref{fig:training}), and fine-tune the two jointly, yielding \texttt{Predictor}$_{\texttt{FT}}$. A symmetric query-set-to-sentence-set \textbf{InfoNCE} objective aligns each image's query set with the sentence-level \texttt{Y-Encoder} embeddings of its reference explanation. Sentences assigned higher teacher-forced negative log-likelihood by the frozen unconditioned decoder receive greater weight, focusing query learning on evidence that the decoder does not already explain well, while a whole-explanation alignment term preserves the Stage~2 semantic representation. Because this permutation-invariant objective admits an identical-slot fixed point, we discourage collapse through annealed sentence-to-slot routing, sentence-level ownership and load-balancing losses, a squared pairwise-cosine penalty, and slot-specific output projections. Third (Stage 3c), with \texttt{Predictor}$_{\texttt{FT}}$, the query module, and \texttt{Y-Decoder} all locked, a zero-initialized residual cross-attention adapter conditions the decoder's visual tokens on the query vectors, in the spirit of gated cross-attention injection into a frozen decoder~\cite{alayrac2022flamingo}; only this adapter is trained, again with \textbf{cross-entropy} on the reference explanations. We give the training protocol in \cref{sec:implementation} and defer the derivation of the full objectives to Appendix~\ref{appendix:training}.

\subsection{Inference}
\label{subsec:method-inference}
 
At inference (\cref{fig:teaser}), a single pass through the adapted \texttt{X-Encoder} feeds \texttt{Classifier}, which reads the verdict $\hat{\ell}$ directly from its classification logits at no decoding cost, and \texttt{Predictor}$_{\texttt{FT}}$ with its query module, which produces the query vectors. \texttt{Y-Encoder} plays no role at inference and is not loaded, since its embedding space is only needed to supervise the predictor during training. The explanation is generated only on demand: the adapter conditions \texttt{Y-Decoder}'s visual tokens on the query vectors, and the image together with the verdict-bearing prompt is decoded into natural language forensic evidence supporting the verdict. The verdict therefore requires no language-model decoding at all, and every base model remains reusable exactly as released, against the billions of parameters fine-tuned end-to-end by MLLM-based systems~\cite{xu2024fakeshield, huang2024sida, li2024forgerygpt, sun2024forgerysleuth, kang2025legionlearninggroundexplain}.
\section{Implementation of \ours}
\label{sec:implementation}
 
\subsection{Model Architecture}
\label{subsec:implementation-architecture}
 
\noindent\textbf{\texttt{X-Encoder}.} We instantiate $f_x$ as DINOv3 ViT-L/16~\cite{simeoni2025dinov3} at $384\times384$ resolution. Its pretrained weights remain frozen, and LoRA parameters~\cite{hu2021lora} of rank $32$ and scale $64$ are inserted into the fused query-key-value projection of each attention block for Stage 1 following~\cite{qu2026dinomac}. We read the last four transformer blocks with their output normalization applied; at each block, the mean-pooled patch tokens, \texttt{[cls]} token, and four register tokens are concatenated, and the four block representations form a sequence of four visual tokens.

\noindent\textbf{\texttt{Classifier}.} $c_\phi$ projects each of the four \texttt{X-Encoder} readouts to a common width with its own linear layer, prepends a learned global token, and applies a four-layer pre-norm transformer with eight attention heads. A linear head from the global-token position produces the two authenticity logits from which the verdict $\hat{\ell}$ is read.
 
\noindent\textbf{\texttt{Predictor}.} $g_\theta$ follows the same configuration as \texttt{Classifier} with independently parameterized weights. Its global token representation is projected to an $\ell_2$-normalized prediction $\hat{\mathbf{s}}_y$ in the semantic embedding space. It is trained from scratch in Stage 2, and fine-tuned jointly with the query module in Stage 3b to give \texttt{Predictor}$_{\texttt{FT}}$.
 
\noindent\textbf{\texttt{Y-Encoder}.} We instantiate $f_y$ as a frozen E5-large-v2~\cite{wang2022e5}, producing normalized text embeddings. Whole-explanation embeddings supervise the global \texttt{Predictor} prediction in Stage 2 and the retention term of Stage 3b, while sentence-level embeddings supervise the query module in Stage 3b; \texttt{Y-Encoder} only precomputes these targets and is discarded at inference.
 
\noindent\textbf{Query module.} Four learned slot vectors cross-attend to a memory of \texttt{Predictor}$_{\texttt{FT}}$'s global hidden state and its projected per-block visual tokens, producing four normalized query vectors $\mathbf{q}_1, \ldots, \mathbf{q}_4$. The slots are exchangeable and receive no predefined forensic category or aspect label.
 
\noindent\textbf{\texttt{Y-Decoder}.} We use $d_\psi$ as Qwen2.5-VL-3B-Instruct~\cite{bai2025qwen25vl}. The vision encoder remains frozen, while a language-side LoRA adapter~\cite{hu2021lora} of rank $16$ is applied to the attention and MLP projections and trained in Stage 3a; the detector verdict is written into the prompt as a word. For query conditioning, a residual cross-attention adapter~\cite{alayrac2022flamingo} projects $d_\psi$'s  visual tokens and $\mathbf{q}_1, \ldots, \mathbf{q}_4$ to a shared width, using the visual tokens as queries and the query vectors as keys and values; its output is projected back to the visual token width and added residually, with the output projection initialized to zero so the decoder is unchanged before adapter training. Only this adapter is trained in Stage 3c; the decoder weights and its LoRA remain locked.
 
\subsection{Three-Stage Training}
\label{subsec:implementation-training}
 
\noindent\textbf{Stage 1 (Verdict Prediction).} We train the \texttt{X-Encoder} LoRA parameters, the \texttt{Classifier} trunk, and its classification readout using \textbf{cross-entropy} loss, with class balancing by undersampling. We use AdamW~\cite{loshchilov2019adamw} with learning rate $10^{-4}$, weight decay $10^{-2}$, and a $512$-step linear warm-up, physical batch size $16$ with four-step gradient accumulation for an effective batch size of $64$. Stage~1 runs for a single epoch over the training split; checkpoints are evaluated at a fixed update interval on the external Chameleon validation set, and the checkpoint with the highest validation AUC is selected (\cf \cref{subsec:ablation-adaptation}). The resulting \texttt{X-Encoder} and \texttt{Classifier} are then locked.
 
\noindent\textbf{Stage 2 (Joint Embedding Prediction).} Using features from the selected \texttt{X-Encoder}, we train \texttt{Predictor} with \textbf{InfoNCE} loss~\cite{oord2018cpc} at temperature $\tau=0.05$ against whole-explanation \texttt{Y-Encoder} embeddings. We use AdamW with sharpness-aware minimization~\cite{foret2021sam}, learning rate $3\times10^{-4}$, weight decay $0.05$, a $500$-step warm-up, and cosine decay, with batch size $2048$ for $24$ epochs.
 
\noindent\textbf{Stage 3a (Decoder Adaptation).} We train the language-side LoRA of \texttt{Y-Decoder} with next-token \textbf{cross-entropy} loss, conditioned on the image and the locked verdict and with no conditioning from \texttt{Predictor}. We use AdamW with learning rate $10^{-4}$, weight decay $10^{-2}$, and cosine decay, with class balancing by undersampling, physical batch size $4$ and eight-step gradient accumulation for an effective batch size of $32$, for one epoch.
 
\noindent\textbf{Stage 3b (Relevance-Weighted Query Training).} We jointly optimize the Stage 2 \texttt{Predictor} and the query module with a symmetric set-to-sentence \textbf{InfoNCE} objective, using slot-aggregation temperature $\tau_s = 0.05$ and contrastive temperature $\tau_c = 0.05$, a whole-explanation global alignment term, and a diversity penalty over the four slots, giving \texttt{Predictor}$_{\texttt{FT}}$. Sentence weights are derived from the unconditioned decoder's per-sentence negative log-likelihood (NLL), so that sentences it already explains poorly receive greater weight. We use AdamW with learning rate $3\times10^{-4}$, weight decay $10^{-2}$, cosine decay, batch size $128$ without gradient accumulation, and class-balanced cycling for $15$ epochs.

\noindent\textbf{Stage 3c (Residual-Adapter Training).} With \texttt{Predictor}$_{\texttt{FT}}$, the query module, and \texttt{Y-Decoder} and its LoRA all locked, we train only the residual cross-attention adapter with next-token \textbf{cross-entropy} loss. We use AdamW with learning rate $10^{-4}$, weight decay $10^{-2}$, and cosine decay, with class balancing by undersampling, physical batch size $4$, eight-step gradient accumulation, and effective batch size $32$, for one epoch. Complete settings for all five optimization steps are collected in Appendix~\ref{appendix:training}.

\noindent\textbf{Cost.} \ours trains 89.26M parameters in total, 28.35M of which define the verdict, against roughly 3.65B frozen weights that are not modified. The verdict itself requires a single \texttt{X-Encoder} pass with no language-model decoding. More in Appendices~\ref{appendix:components} and~\ref{appendix:training}.
\section{Experiments}
\label{sec:experiments}

\begin{figure*}[t]
\centering
\small
\setlength{\tabcolsep}{4pt}
\renewcommand{\arraystretch}{1.1}
\begin{tabular}{@{}>{\centering\arraybackslash}m{0.16\linewidth} m{0.40\linewidth} m{0.40\linewidth}@{}}
  \multicolumn{1}{c}{\footnotesize Image}
  & \multicolumn{1}{c}{\footnotesize \ours\ (ours)}
  & \multicolumn{1}{c}{\footnotesize Ground Truth} \\[4pt]
  \includegraphics[width=\linewidth,height=2cm,keepaspectratio]{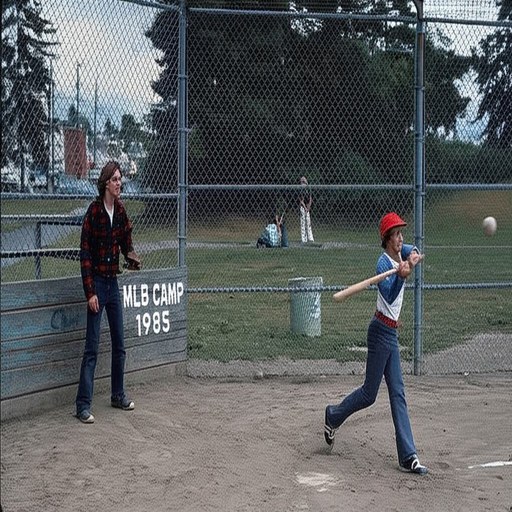}
  & \scriptsize The image contains several visual inconsistencies that suggest manipulation. Most notably, the `MLB CAMP 1985' text on the wooden fence appears unnaturally sharp and flat compared to the weathered texture of the wood it is painted on; the letters do not follow the slight curves or grain of the wood surface. Additionally, there is a noticeable smudge and loss of detail in the area directly behind the fence where the background elements seem to have been blurred or erased, leaving a patchy, unnatural texture in the grass and trees.
  & \scriptsize The image contains several visual inconsistencies that suggest digital manipulation. Most notably, the white text `MLB CAMP 1985' on the wooden fence appears unnaturally sharp and bright compared to the grainy, weathered texture of the wood behind it. The letters do not follow the physical gaps or grain of the wooden planks, making them look like a flat digital overlay. Additionally, there is a faint, blurry blue smudge visible just to the left of the text that doesn't correspond to any physical object in the scene. Furthermore, the batter's hands and the handle of the baseball bat show significant distortion, with the fingers appearing to blend into the wood of the bat in a smeared, unnatural way. \\
  \addlinespace[2pt]
  \includegraphics[width=\linewidth,height=2cm,keepaspectratio]{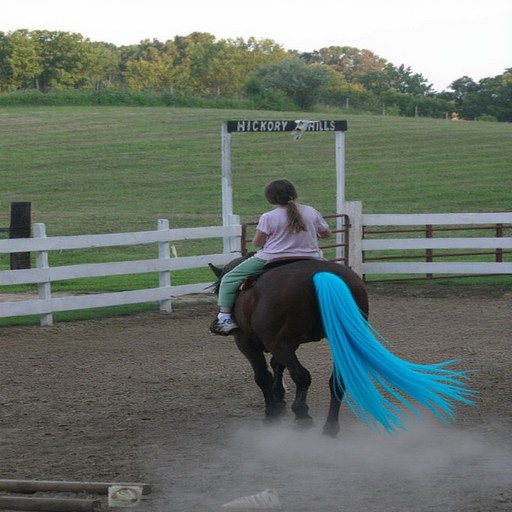}
  & \scriptsize The image contains a significant visual anomaly where the horse's tail appears as a bright, neon blue color that lacks any natural texture or lighting consistency with the rest of the animal. This vibrant blue hue is unnaturally saturated and does not match the muted, earthy tones of the horse's coat or the surrounding environment. Furthermore, the edges of this blue tail are overly sharp and lack the motion blur seen in the horse's legs and the dust kicked up by its hooves, making it look like a flat digital overlay rather than a physical part of the scene.
  & \scriptsize The image contains a significant visual anomaly in the horse's tail, which appears as a vibrant, neon blue color that is inconsistent with the natural brown and earthy tones of the rest of the scene. Furthermore, the texture of the blue hair is overly smooth and lacks the fine detail and motion blur seen in the dust kicked up by the horse's hooves, making it look like a digital overlay rather than a physical part of the animal. \\
\end{tabular}
\caption{\textbf{Qualitative explanations produced by \ours.} For each detected fake: input image (\textbf{left}), the explanation produced by the guided \texttt{Y-Decoder} (\textbf{center}), and the ground-truth forensic annotation (\textbf{right}).}
\label{fig:qual_examples}
\end{figure*}

\begin{table*}
\centering
\small
\setlength{\tabcolsep}{1.3pt}
\caption{\textbf{Detection performance on unseen generators.} AUC ($\times100$) on 14 unseen generators. Best per column in \textbf{bold}, second best \underline{underlined}, with ties bolded jointly. The \mbox{XPV} ows indicate retraining on XPlainVerse.}
\label{tab:detection}
\begin{tabular}{lccccccccccccccc}
\toprule
&
\multicolumn{3}{c}{\textbf{Stable Diffusion}} &
\multicolumn{3}{c}{\textbf{Midjourney}} &
\multicolumn{2}{c}{\textbf{DALL-E}} &
\multicolumn{2}{c}{\textbf{GPT}} &
\multicolumn{4}{c}{\textbf{Other}} &
 \\
\cmidrule(lr){2-4}
\cmidrule(lr){5-7}
\cmidrule(lr){8-9}
\cmidrule(lr){10-11}
\cmidrule(lr){12-15}
\textbf{Approach} & SD2 & SDXL & SD3 & MJv5 & MJv6.1 & MJ7 &
DALLE2 & DALLE3 & GPT-I2 & GPT-I1.5 &
Flux & GigaGAN & Firefly & NB-Pro & \textbf{Mean} \\
\midrule
DeFake~\cite{sha2023fake}
& 76.3 & 84.5 & \underline{95.6} & 83.5 & 86.4 & 81.9
& 68.1 & 94.8 & \underline{82.6} & \underline{80.0}
& \underline{95.5} & 58.9 & 84.5 & 71.1 & 81.7 \\

PatchCr.~\cite{zhong2024patchcraft}
& 95.5 & 96.3 & 32.5 & 78.0 & 95.7 & \underline{94.8}
& 80.0 & 28.5 & 59.5 & 78.7
& 67.2 & 86.9 & 78.2 & 59.7 & 73.7 \\

RINE~\cite{koutlis2024leveraging}
& 83.6 & 86.5 & 30.7 & 75.7 & 67.7 & 63.7
& 75.6 & 72.3 & 29.4 & 42.7
& 68.1 & 93.4 & 70.7 & 33.5 & 63.8 \\

SPAI~\cite{karageorgiou2025spai}
& 93.3 & 94.0 & 61.2 & 97.4 & 80.4 & 83.7
& 83.2 & 81.1 & 63.7 & 78.0
& 80.0 & 92.9 & 87.6 & 60.8 & 81.2 \\

B-Free~\cite{guillaro2024biasfree}
& \textbf{100.0} & \textbf{100.0} & 92.7
& \textbf{99.9} & 96.2 & 92.7
& 98.8 & \underline{99.5}
& 73.1 & 66.8
& 93.9 & \underline{96.4} & \textbf{99.9} & \underline{96.7}
& 93.3 \\

FakeVLM~\cite{wen2025fakevlm}
& 64.8 & 58.4 & 33.1 & 57.0 & 52.5 & 49.5
& 64.8 & 37.1 & 36.6 & 46.9
& 32.7 & 83.1 & 40.6 & 56.4 & 51.0 \\

SIDA~\cite{huang2024sida}
& 61.2 & 87.5 & \textbf{96.6} & 94.6 & 96.1 & 95.5
& 61.1 & \underline{99.7} & 82.3 & 85.5
& 99.3 & 66.4 & \underline{90.6} & 91.1 & 86.3 \\

SPAI$_{\text{XPV}}$\cite{karageorgiou2025spai}
& 77.7 & 92.0 & 73.2 & 85.4 & 93.8 & 79.0
& 71.9 & 90.6 & 62.1 & 73.0
& 91.9 & 89.9 & 84.7 & 94.9 & 82.9 \\

RINE$_{\text{XPV}}$\cite{koutlis2024leveraging}
& 83.8 & 98.5 & 79.5 & 95.1 & 95.2 & 86.3
& 89.5 & 98.0 & 74.8 & 76.1
& 93.8 & 80.1 & 86.0 & 93.8 & 87.9 \\

B-Free$_{\text{XPV}}$\cite{guillaro2024biasfree}
& 90.8 & 92.3 & 94.5 & 99.7 & 94.3 & 92.6
& 99.5 & 99.1 & 78.9 & 83.2
& 98.7 & 80.1 & 82.1 & 92.9 & 91.4 \\

DINOMAC$_{\text{XPV}}$\cite{qu2026dinomac}
& 99.1 & \underline{99.9} & 94.0
& 99.1 & \underline{98.3} & 78.9
& \underline{99.7} & \textbf{100.0}
& 81.1 & 81.3
& \underline{99.8} & \textbf{99.8} & 86.4
& 94.2 & \underline{93.7} \\

\ours\ (ours)
& \underline{99.3} & \textbf{100.0} & \textbf{96.1}
& \underline{99.8} & \textbf{99.0} & \textbf{96.6}
& \textbf{99.8} & \textbf{100.0}
& \textbf{90.7} & \textbf{94.9}
& \textbf{99.9} & \textbf{99.8} & \underline{90.0}
& \textbf{97.3} & \textbf{97.4} \\
\bottomrule
\end{tabular}
\end{table*}

\begin{table}
\centering
\small
\setlength{\tabcolsep}{1.7pt}
\caption{\textbf{Explanation performance comparison.} ROUGE-L (R-L), CSS~\cite{kang2025legionlearninggroundexplain}, and BERTScore F1~\cite{zhang2020bertscore:} ($\times100$). $^\dagger$No public LEGION detection checkpoint is available, entries are from~\cite{kang2025legionlearninggroundexplain}.}
\label{tab:explanation}
\begin{tabular}{lcccccc}
\toprule
& \multicolumn{3}{c}{\textbf{XPlainVerse}}
& \multicolumn{3}{c}{\textbf{LOKI}} \\
\cmidrule(lr){2-4} \cmidrule(lr){5-7}
\textbf{Approach}
& R-L & CSS & BERTSc.
& R-L & CSS & BERTSc. \\
\midrule
SIDA~\cite{huang2024sida}
& 15.7 & 40.4 & 52.3
& 19.2 & 40.9 & 53.4 \\

FakeVLM~\cite{wen2025fakevlm}
& 17.0 & 47.1 & 57.0
& 17.4 & 51.4 & 55.8 \\

LEGION~\cite{kang2025legionlearninggroundexplain}$^\dagger$
& n/a & n/a & n/a
& 18.5 & 45.9 & n/a \\

\ours (Stage 3a only)
& 25.5 & 59.9 & 65.3
& 17.1 & 48.9 & 52.1 \\
\ours (proposed)
& \textbf{27.5} & \textbf{62.2} & \textbf{69.8}
& \textbf{20.1} & \textbf{52.0} & \textbf{56.1} \\
\bottomrule
\end{tabular}
\end{table}

\subsection{Experimental Setup}
\label{subsec:experiments-setup}

\noindent\textbf{Metrics.}
We report AUC for detection. Explanation quality is measured with ROUGE-L, CSS~\cite{kang2025legionlearninggroundexplain}, and BERTScore F1~\cite{zhang2020bertscore:} against the reference explanation.

\noindent\textbf{Datasets.}
\ours is trained on the XPlainVerse (\mbox{XPV})~\cite{narang2026xplainverse} training split, derived from MultiFakeVerse~\cite{gupta2025multifakeverse}. \textbf{Detection} transfer is evaluated on a benchmark assembled from two sources. From the any-resolution protocol of SPAI~\cite{karageorgiou2025spai}, we retain the 10 generators whose outputs exceed 0.5 megapixels, spanning latent diffusion~\cite{rombach2022high,podell2024sdxl,esser2024scaling}, flow-based~\cite{flux2024}, GAN~\cite{kang2023gigagan}, and commercial~\cite{midjourney,ramesh2022hierarchical,betker2023improving,firefly} models. We additionally use four frontier generators from the held-out out-of-distribution split of OpenFake~\cite{livernoche2025openfake}, which also shifts the real image distribution. \textbf{Explanation} is evaluated on the held-out XPlainVerse split and LOKI~\cite{ye2024loki}. LOKI is outside the XPlainVerse training distribution and therefore provides an external test of whether forensic explanations transfer robustly to an independent set. More in Appendix~\ref{appendix:evaluation}.

\noindent\textbf{Baselines.}
For detection, we compare with DeFake~\cite{sha2023fake}, PatchCraft~\cite{zhong2024patchcraft}, RINE~\cite{koutlis2024leveraging}, SPAI~\cite{karageorgiou2025spai}, and B-Free~\cite{guillaro2024biasfree}. We also retrain SPAI, RINE, B-Free and DINO-MAC on XPlainVerse. The latter retrained baselines use the same external Chameleon validation protocol as \ours. For explanation, we compare with SIDA~\cite{huang2024sida}, FakeVLM~\cite{wen2025fakevlm}, and LEGION~\cite{kang2025legionlearninggroundexplain}. Unless stated otherwise, baseline results are recomputed from public checkpoints under the same evaluation protocol. More in Appendix~\ref{appendix:results}.

\subsection{Comparison with State of the Art}
\label{subsec:experiments-sota}

\begin{figure*}[!t]
\centering
\includegraphics[width=\linewidth]{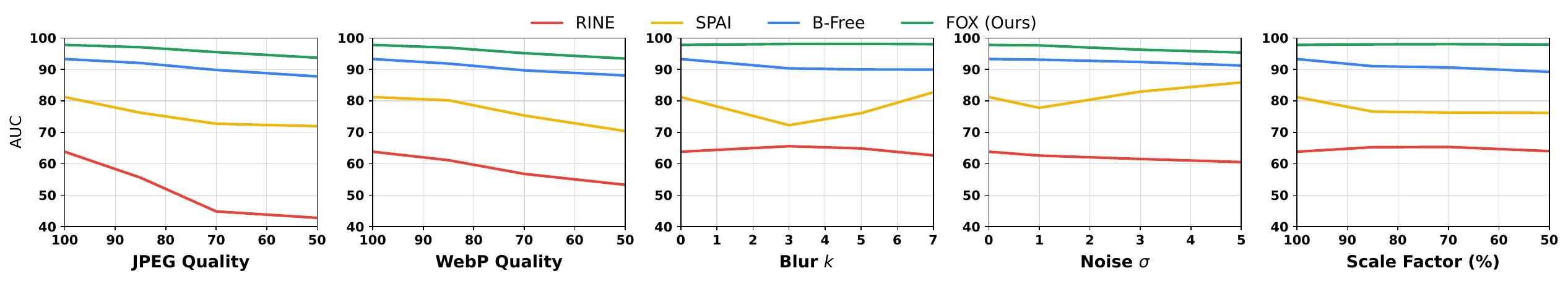}
\caption{\textbf{Robustness to common perturbations.} AUC ($\times100$) of \ours and representative baselines under increasing severity of JPEG compression, WebP compression, downscaling, additive Gaussian noise, and blur.}
\label{fig:robustness}
\end{figure*}

\noindent\textbf{Detection performance on unseen generators.}
\Cref{tab:detection} shows strong transfer across all generator families: \ours improves mean AUC by \textbf{+3.7 points} over the runner-up, DINO-MAC~\cite{qu2026dinomac}. The \mbox{XPV} controls show that this advantage cannot be attributed solely to training on XPlainVerse.

\noindent\textbf{Explanation quality.}
\Cref{tab:explanation} shows that \ours achieves the strongest similarity on both XPlainVerse and the external LOKI. To isolate the detector-derived guidance, we compare against a Stage~3a variant that receives only the verdict at inference. The full pathway improves every metric on both datasets, with gains on LOKI at least as large as in domain, while the variant falls below FakeVLM on all three LOKI metrics. Therefore, the predicted caption-space representation (and not the decoder fine-tuning)  is the part that carries explanation quality to an unseen forensic distribution. The qualitative examples in \cref{fig:qual_examples} show that the decoder captures the principal forensic evidence in the references.

\subsection{Robustness to Common Perturbations}
\label{subsec:experiments-robustness}

\Cref{fig:robustness} shows that \ours is substantially more robust than the compared detectors under compression, downscaling, noise, and blur. DINOv3~\cite{simeoni2025dinov3} inherits invariances from view-consistency pretraining and augmentations that include scaling and blur~\cite{caron2021emerging,oquab2023dinov2}, whereas forensic methods that rely on high-frequency generator traces are particularly vulnerable to such transformations~\cite{corvi2023intriguingproperties}. We therefore view preservation of these pretrained invariances through brief adaptation as a plausible explanation of the observed robustness.



\section{Ablation Study of \ours}
\label{sec:ablations}

All ablations use the same DINOv3 backbone, training data, and evaluation protocol, isolating the effect of each architectural or optimization choice while controlling all remaining components. We report mean unseen-generator AUC for detection and BERTScore F1 on the held-out validation split for explanation quality. More in Appendix~\ref{appendix:ablations}.

\subsection{Backbone Choice}
\label{subsec:ablation-backbone}

\emph{Which representation should perform forensic detection?} Prior work has predominantly relied on language-aligned representations~\cite{ojha2023towards, cozzolino2024raising, koutlis2024leveraging}, while recent forensic systems increasingly exploit pure, self-supervised~\cite{giakoumoglou2024review} vision DINO features~\cite{huang2025rethinking, qu2026dinomac}. \Cref{tab:backbone} compares these families under the same frozen readout and shows that view-consistency representations transfer most reliably across unseen generators, with DINOv3 consistently being the strongest. This supports our use of DINOv3 as the specialist visual encoder while reserving language-aligned models for explanation.

\begin{table}[!htbp]
\vspace{-6pt}
\centering
\small
\setlength{\tabcolsep}{2.5pt}
\caption{\textbf{Ablation on backbone choice.} Frozen backbones share the same downstream readout and are evaluated on five unseen GenImage generators.}
\label{tab:backbone}
\begin{tabular}{lcccccc}
\toprule
\textbf{Backbone}
& \textbf{SD1.5}
& \textbf{BigGAN}
& \textbf{VQDM}
& \textbf{MJv4}
& \textbf{Wuk}
& \textbf{Mean} \\
\midrule
\multicolumn{7}{l}{\textit{Language-supervised}} \\
CLIP~\cite{radford2021learning}
& 88.6 & 90.9 & 93.2 & 81.9 & 86.1 & 88.1 \\
SigLIP~\cite{zhai2023sigmoid}
& 89.7 & 86.8 & 84.4 & 80.3 & 91.5 & 86.5 \\
AIMv2~\cite{fini2025aimv2}
& 84.9 & 86.6 & 82.8 & 85.4 & 88.5 & 85.6 \\
\midrule
\multicolumn{7}{l}{\textit{Pure-vision, generative or predictive}} \\
MAE~\cite{he2022mae}
& 83.1 & 84.2 & 87.6 & 85.5 & 83.1 & 84.7 \\
I-JEPA~\cite{assran2023ijepa}
& 78.4 & 61.4 & 62.2 & 81.9 & 75.7 & 71.9 \\
V-JEPA 2~\cite{assran2025vjepa2}
& 71.7 & 57.8 & 54.9 & 87.8 & 73.4 & 69.1 \\
\midrule
\multicolumn{7}{l}{\textit{Pure-vision, view-consistency}} \\
DINOv2~\cite{oquab2023dinov2}
& 94.8 & 96.9 & 96.0 & 89.7 & 96.1 & 94.7 \\
DINOv3~\cite{simeoni2025dinov3}
& \cellcolor{gray!10}\textbf{99.3}
& \cellcolor{gray!10}\textbf{99.2}
& \cellcolor{gray!10}\textbf{99.2}
& \cellcolor{gray!10}\textbf{96.9}
& \cellcolor{gray!10}\textbf{99.5}
& \cellcolor{gray!10}\textbf{98.8} \\
\bottomrule
\end{tabular}
\vspace{-8pt}
\end{table}

\subsection{Backbone Adaptation}
\label{subsec:ablation-adaptation}

\emph{How should the selected backbone be adapted?} Prior work finds strong transfer from frozen probes~\cite{huang2025rethinking, ojha2023towards} and parameter-efficient adaptation~\cite{qu2026dinomac, liu2024forgery}, leaving selection unclear. \Cref{tab:adaptation} shows that XPV validation AUC is nearly saturated across adaptation strategies, despite substantial differences in cross-generator transfer, making it unreliable for checkpoint selection. Selecting the LoRA checkpoint using external Chameleon validation instead yields the strongest transfer to the GenImage generators. 

\begin{table}[!htbp]
\vspace{-6pt}
\centering
\small
\setlength{\tabcolsep}{3pt}
\caption{\textbf{Ablation on backbone adaptation.} We compare freezing, end-of-epoch LoRA}
\label{tab:adaptation}
\begin{tabular}{lccc}
\toprule
\textbf{DINOv3}
& \textbf{XPV}
& \textbf{Cham.}
& \textbf{GenImage} \\
\midrule
Frozen
& 99.4 & 79.9 & 98.8 \\
LoRA
& 99.9 & 83.6 & 97.6 \\
LoRA + int. val. (XPV)
& 99.9 & 82.6 & 95.0 \\
LoRA + ext. val. (Cham.) (ours)
& \cellcolor{gray!10}99.9
& \cellcolor{gray!10}\textbf{97.2}
& \cellcolor{gray!10}\textbf{99.7} \\
\bottomrule
\end{tabular}
\vspace{-8pt}
\end{table}

\subsection{End-to-End versus Decoupled Training}
\label{subsec:ablation-e2e}

\emph{Should detection and explanation be optimized within one pipeline?}
Existing explainable detectors commonly train multimodal models to predict authenticity and generate its explanation jointly~\cite{xu2024fakeshield,huang2024sida, li2024forgerygpt, sun2024forgerysleuth}. \Cref{tab:ablation_decoupling} compares this design with \ours, where the detector is locked before explanation training. Decoupling provides a stronger combination of detection transfer and explanation quality. The shared-trunk variants further expose a checkpoint trade-off: stopping earlier favours detection, whereas continuing optimization favours explanation alignment. Disjoint trunks avoid this competition and allow each objective to converge independently, supporting the central design choice of separating specialist perception from language generation.

\begin{table}[!htbp]
\vspace{-6pt}
\centering
\small
\setlength{\tabcolsep}{1.5pt}
\caption{\textbf{Ablation on training design.} Joint versus separate optimization and shared versus disjoint trunks.}
\label{tab:ablation_decoupling}
\begin{tabular}{lcc}
\toprule
\textbf{Variant}
& \textbf{AUC (Unseen)}
& \textbf{BERTSc.} \\
\midrule
\multicolumn{3}{l}{\textit{Training organization}} \\
Single end-to-end pipeline
& 93.4 & 68.9 \\
Decoupled detector + explainer (ours)
& \cellcolor{gray!10}\textbf{97.4}
& \cellcolor{gray!10}\textbf{72.1} \\
\midrule
\multicolumn{3}{l}{\textit{Representation sharing}} \\
Shared trunk, early stop
& 96.9 & 68.9 \\
Shared trunk, late stop
& 94.5 & 69.5 \\
Disjoint trunks (ours)
& \cellcolor{gray!10}\textbf{97.4}
& \cellcolor{gray!10}\textbf{72.1} \\
\bottomrule
\end{tabular}
\vspace{-8pt}
\end{table}

\subsection{Query Training Components}
\label{subsec:ablation-query-training}

\emph{Which components prevent query collapse and improve explanation quality?} \Cref{tab:ablation_query_training} isolates collapse resistance, global retention, and relevance weighting (of Stage~3b). Collapse resistance reduces query similarity  and improves explanation quality. Global retention raises explanation quality, while NLL weighting outperforms uniform weighting and produces more distinct queries.

\begin{table}[!htbp]
\vspace{-6pt}
\centering
\small
\setlength{\tabcolsep}{3pt}
\caption{\textbf{Ablation on query training.}. Cosine similarity is the mean pairwise similarity between query vectors ($\times100$); lower values indicate less query collapse.}
\label{tab:ablation_query_training}
\begin{tabular}{lcccc}
\toprule
\multirow{2}{*}{\textbf{Weighting}}
& \textbf{Global}
& \textbf{Collapse}
& \multirow{2}{*}{\textbf{BERTSc.}}
& \multirow{2}{*}{\textbf{Cos. sim.}$\downarrow$} \\
& \textbf{retention}
& \textbf{resistance}
& & \\
\midrule
NLL     & No  & No  & 68.5 & 99.9 \\
NLL     & No  & Yes & 69.3 & 35.7 \\
Uniform & Yes & Yes & 70.2 & 68.4 \\
NLL     & \cellcolor{gray!10}Yes
& \cellcolor{gray!10}Yes
& \cellcolor{gray!10}\textbf{72.1}
& \cellcolor{gray!10}\textbf{31.1} \\
\bottomrule
\end{tabular}
\vspace{-8pt}
\end{table}

\subsection{VLM Backbone and Explanation Pathway}
\label{subsec:ablation-explanation}

\emph{Does stronger explanation require a stronger VLM or better forensic guidance?} Explainable detectors commonly invest substantial capacity in the multimodal model itself~\cite{xu2024fakeshield, huang2024sida, sun2024forgerysleuth}, whereas \ours uses the VLM only for verbalization. \Cref{tab:ablation_explanation} shows that the selected compact Qwen backbone is competitive before adaptation, while the full pathway becomes strongest only with detector-derived query conditioning. The weak improvement from a query-free adapter isolates the benefit of the predicted forensic representation.

\begin{table}[!htbp]
\vspace{-6pt}
\centering
\small
\setlength{\tabcolsep}{3pt}
\caption{\textbf{Ablation on the explanation.} \textbf{Top:} VLM backbone. \textbf{Bottom:} progressive conditioning with Qwen2.5-VL-3B fixed.}
\label{tab:ablation_explanation}
\begin{tabular}{lc}
\toprule
\textbf{Stage 3: Configuration}
& \textbf{BERTSc.} \\
\midrule
\multicolumn{2}{l}{\textit{Stage 3a: image + verdict: VLM ablation}} \\
PaliGemma-2-3B Mix
& 66.3 \\
InternVL2.5-4B
& 66.2 \\
Qwen2.5-VL-3B (selected)
& 69.1 \\
\midrule
\multicolumn{2}{l}{\textit{Qwen2.5-VL-3B explanation pathway}} \\
Stage 3a + 3c, no queries
& 69.3 \\
Stage 3a + 3b + 3c (ours)
& \cellcolor{gray!10}\textbf{72.1} \\
\bottomrule
\end{tabular}
\vspace{-8pt}
\end{table}

\subsection{Query Set Size}
\label{subsec:ablation-queries}

\emph{How much detector-derived guidance should the VLM receive?} A single query may compress forensic cues into one representation, whereas excessive capacity may add redundant conditioning. \Cref{tab:ablation_queries} shows that four exchangeable queries provide the best balance. Shuffling these queries across images while preserving their number and marginal distribution degrades explanation quality, showing that their benefit comes from image-specific forensic content.

\begin{table}[!htbp]
\vspace{-6pt}
\centering
\small
\setlength{\tabcolsep}{3pt}
\caption{\textbf{Ablation on query set size.} We vary the number of slots while fixing the remaining explanation pathway.}
\label{tab:ablation_queries}
\begin{tabular}{lc}
\toprule
\textbf{Stage 3b: Query slots}
& \textbf{BERTSc.} \\
\midrule
$1$ slot
& 69.3 \\
$2$ slots
& 69.9 \\
$4$ slots (ours)
& \cellcolor{gray!10}\textbf{72.1} \\
$8$ slots
& 70.1 \\
$4$ slots, shuffled across images
& 68.9 \\
\bottomrule
\end{tabular}
\vspace{-8pt}
\end{table}
\section{Conclusion}
\label{sec:conclusion}

We presented \ours, which decouples reliable detection from forensic explanation. A briefly adapted DINOv3 detector transfers strongly across unseen generators and remains robust to common perturbations. Once locked, a joint-embedding predictor maps its visual evidence into a frozen text space that guides a lightweight VLM. Explanation training therefore cannot alter the verdict, and still benefits from the specialist detector's representation. Our results demonstrate that strong detection and explanation can work synergetically: the detector decides, and language is limited to explaining the decision.

\section*{Acknowledgements}

This work was supported by the Horizon Europe projects ELLIOT (Grant Agreement No. 101214398) and AI-CODE (Grant Agreement No. 101135437).

{
    \small
    \bibliographystyle{ieeenat_fullname}
    \bibliography{main}
}

\onecolumn
\appendix
\renewcommand{\thesubsection}{\Alph{section}.\arabic{subsection}}
\renewcommand{\thesubsubsection}{\Alph{section}.\arabic{subsection}.\arabic{subsubsection}}
\renewcommand{\thefigure}{\Alph{section}.\arabic{figure}}
\renewcommand{\thetable}{\Alph{section}.\arabic{table}}
\renewcommand{\theequation}{\Alph{section}.\arabic{equation}}
\counterwithin*{figure}{section}
\counterwithin*{table}{section}
\counterwithin*{equation}{section}
\setlength{\parindent}{0pt}
\setlength{\parskip}{0.5\baselineskip plus 2pt minus 1pt}
\setlength{\parfillskip}{0pt plus 1fil}
\makeatletter
\def\@listI{\leftmargin\leftmargini \topsep\z@ \parsep\parskip \itemsep\z@}\let\@listi\@listI \@listi
\def\@listii{\leftmargin\leftmarginii \labelwidth\leftmarginii\advance\labelwidth-\labelsep \topsep\z@ \parsep\parskip \itemsep\z@}
\def\@listiii{\leftmargin\leftmarginiii \labelwidth\leftmarginiii\advance\labelwidth-\labelsep \topsep\z@ \parsep\parskip \itemsep\z@}
\partopsep=\z@
\patchcmd\@startsection{\addvspace\@tempskipa}{\advance\@tempskipa-\parskip\addvspace\@tempskipa}{}{}
\patchcmd\@xsect{\vskip\@tempskipa}{\advance\@tempskipa-\parskip\vskip\@tempskipa}{}{}
\makeatother
\begin{center}
  \Large\textbf{\thetitle}\\[0.5em]
  Supplementary Material\\[1em]
\end{center}

\section{Extended Related Work}
\label{appendix:related}

\subsection{AI-Generated Image Detection}
\label{appendix:related-aigi}

A \textbf{first} family targets low-level generation artifacts. CNNSpot~\cite{wang2020cnn} showed that a classifier trained on ProGAN images generalizes to unseen GANs with careful augmentation, and subsequent work located traces in the frequency domain~\cite{durall2020watch, frank2020leveraging, zhang2019detecting, corvi2023detection, bammey2023synthbuster, li2024masksim, doloriel2024frequency, tan2024frequency}, in gradient statistics~\cite{tan2023learning}, in upsampling-induced pixel relationships~\cite{tan2024npr}, in co-occurrence and correlation cues fused across streams~\cite{ju2022fusing}, in global texture statistics~\cite{liu2020gramnet}, and in texture-rich patches~\cite{zhong2024patchcraft}. SPAI~\cite{karageorgiou2025spai} reframes detection as one-class, modeling the spectral distribution of real images at arbitrary resolution and flagging synthetic ones as outliers. These cues sit close to image formation, which is their appeal, but the fingerprint a generator leaves is a property of its weights and sampling procedure~\cite{marra2019gans, yu2019attributing}, so it shifts with every release and is attenuated by the resizing and compression that images undergo online~\cite{corvi2023intriguingproperties}. A \textbf{second} family exploits semantics by probing frozen encoders: UnivFD~\cite{ojha2023towards} linearly probes CLIP~\cite{radford2021learning} embeddings, later refined through intermediate-block representations~\cite{koutlis2024leveraging}, training recipes~\cite{cozzolino2024raising}, forgery-aware adaptation~\cite{liu2024forgery}, hybrid low-level and semantic streams~\cite{yan2024sanity}, and text-conditioned variants~\cite{sha2023fake}. A \textbf{third} family measures how faithfully a generative model reconstructs an image, via diffusion reconstruction error~\cite{wang2023dire}, refined inversions~\cite{ricker2024aeroblade, cazenavette2024fakeinversion}, intermediate latents~\cite{luo2024lare}, local reconstruction statistics~\cite{he2024rigid}, and training-free variants~\cite{cozzolino2025zero}, or contrasts reconstructions during training~\cite{chendrct}. A \textbf{fourth}, emerging family recasts detection as an explainable task addressed by multimodal large language models. General-purpose models trail specialist detectors on the bare binary decision~\cite{jia2024can, li2024fakebench, ye2024loki}, which has motivated forensic variants supervised for manipulated faces~\cite{zhang2024common, huang2024ffaa}, expert-feature fusion~\cite{chen2025xdfd}, segmentation and trace encoders~\cite{xu2024fakeshield, huang2024sida, li2024forgerygpt, sun2024forgerysleuth, kang2025legionlearninggroundexplain}, preference optimization~\cite{zhou2025aigiholmes}, and VQA-style prompting~\cite{chang2023antifakeprompt, keita2024bi}, with FakeVLM~\cite{wen2025fakevlm} and SIDA~\cite{huang2024sida} pairing the verdict directly with a natural-language rationale; these systems train billions of parameters for what is fundamentally a perception problem, and because verdict and rationale are decoded from one set of weights, the two objectives trade against each other. A \textbf{fifth} line attributes part of the transfer gap to the training corpus rather than to the detector, either by controlling the biases a detector can absorb from it~\cite{guillaro2024biasfree} or by redesigning the benchmark so that provenance cannot be mistaken for authenticity, as in GenImage~\cite{zhu2023genimage}, the any-resolution protocol of SPAI~\cite{karageorgiou2025spai}, and OpenFake~\cite{livernoche2025openfake}, which holds out the real-image distribution alongside the generators. Across all these families, models are validated in-domain, degrade on unseen generators~\cite{gragnaniello2021gan, chai2020makes, zhu2023genimage}, and, with the exception of the fourth, return a bare score with no account of the evidence behind it. \ours addresses this at the representation level: the verdict function is selected using a labeled external validation distribution and locked before any explanation objective is initialized, so neither further adaptation nor explanation training can specialize it to the training generators.

\subsection{Joint Embedding Predictive Architectures}
\label{appendix:related-jepa}

Unlike generative models that reconstruct raw inputs~\cite{he2022mae, bao2022beit}, a JEPA~\cite{lecun2022path} predicts the \emph{embedding} of a target signal from a related context signal, discarding detail that is unpredictable in principle rather than spending capacity on it. I-JEPA~\cite{assran2023ijepa} instantiated this for images; the principle has been extended to video~\cite{bardes2024vjepa, assran2025vjepa2, murlabadia2026vjepa2_1} and carried into language modeling~\cite{huang2026llmjepa} and vision-language learning~\cite{chen2026vljepa, kuhn2026levljepa}, where the latter predicts continuous text embeddings instead of autoregressively generating tokens, which is the direction closest to \ours because it separates what an utterance means from the surface form that realizes it. A central concern in this line is preventing the target representation from collapsing to a trivial constant, a failure characterized both for joint-embedding methods in general~\cite{jing2022understanding, giakoumoglou2026three} and for JEPAs specifically~\cite{kenneweg2025jepa}, and addressed through momentum target encoders~\cite{grill2020bootstrap, he2020momentum, assran2023ijepa}, self-distillation with stop-gradient~\cite{baevski2022data2vec}, variance-covariance regularization~\cite{bardes2022vicreg}, and, more recently, the provable single-term regularizer of LeJEPA~\cite{balestriero2025lejepa}. \ours sidesteps this machinery entirely: because the \texttt{Y-Encoder} is frozen and never trained, the target space is fixed by construction and cannot collapse, while InfoNCE~\cite{oord2018cpc} penalizes a constant predictor directly. The design also differs from CLIP-style contrastive learning~\cite{radford2021learning}; instead of co-training two encoders for generic image-text matching, we train neither and learn a one-directional mapping from frozen visual features into a fixed caption space, narrowing the objective to the single question of which manipulation explanation fits the image, and its sentence-level extension yields an unordered set of continuous evidence vectors consumed by an otherwise frozen verbalizer~\cite{alayrac2022flamingo}.

\section{Extended Methodology}
\label{appendix:components}

This section specifies the information passed between the five components
of \ours. Training objectives and optimization settings are deferred to
Appendix~\ref{appendix:training}.  Checkpoint identifiers below are given verbatim: \texttt{timm}~\cite{rw2019timm} names resolve through the \texttt{timm} model registry, and \texttt{org/name} paths resolve on the Hugging Face Hub.

\subsection{X-Encoder}
\label{appendix:xencoder}

The \texttt{X-Encoder} $f_x$ is DINOv3 ViT-L/16
\cite{simeoni2025dinov3} (\texttt{vit\_large\_patch16\_dinov3.lvd1689m}), evaluated at $384\times384$ resolution. We read
the normalized outputs of its last four transformer blocks. At block
$l$, the mean-pooled patch tokens, \texttt{[cls]} token, and four register
tokens are concatenated:
\begin{equation}
    \mathbf{v}^{(l)}
    =
    \operatorname{concat}
    \left(
        \operatorname{mean}(\mathbf{P}^{(l)}),
        \mathbf{c}^{(l)},
        \mathbf{r}^{(l)}_1,\ldots,\mathbf{r}^{(l)}_4
    \right).
\end{equation}
Each constituent token has dimension $1024$, giving
$\mathbf{v}^{(l)}\in\mathbb{R}^{6144}$. The complete visual
representation is
\begin{equation}
    \mathbf{s}_x
    =
    \left[
        \mathbf{v}^{(20)},
        \mathbf{v}^{(21)},
        \mathbf{v}^{(22)},
        \mathbf{v}^{(23)}
    \right]
    \in
    \mathbb{R}^{4\times6144}.
    \label{eq:xencoder_representation}
\end{equation}
Thus, the downstream modules receive four compact block-level vectors
rather than the complete sequence of spatial patch tokens.

The released DINOv3 parameters remain frozen. Stage~1 inserts rank-$32$
LoRA parameters with scale $64$ into the fused query--key--value
projections. After Stage~1, these parameters are locked and the same
adapted representation is used by the \texttt{Classifier} and
\texttt{Predictor}.

\subsection{Classifier and Predictor}
\label{appendix:classifier-predictor}

Both modules process the four vectors in
\cref{eq:xencoder_representation} using the same trunk configuration but
independently parameterized weights. Each block vector is projected to
width $512$, a learned global token is prepended, and learned positional
embeddings are added. A four-layer pre-normalized transformer with eight
attention heads processes the resulting sequence.

The \texttt{Classifier} $c_{\phi}$ maps the final global-token representation to two
authenticity logits:
\begin{equation}
    \mathbf{u}
    =
    c_{\phi}(\mathbf{s}_x)
    \in
    \mathbb{R}^{2}.
\end{equation}
Its softmax output defines the detector score, while the verdict is
\begin{equation}
    \hat{\ell}
    =
    \operatorname*{arg\,max}_{j\in\{0,1\}} u_j ,
\end{equation}
where \(j=0\) denotes real and \(j=1\) generated.

The \texttt{Predictor} $g_{\theta}$ instead maps its global-token representation into
the fixed semantic space of the \texttt{Y-Encoder}:
\begin{equation}
    \hat{\mathbf{s}}_y
    =
    \frac{
        g_{\theta}(\mathbf{s}_x)
    }{
        \|g_{\theta}(\mathbf{s}_x)\|_2
    }.
    \label{eq:global_semantic_prediction}
\end{equation}
The \texttt{Predictor} has no classification readout. It is trained after
the detector has been locked, so semantic prediction cannot alter the
verdict function.

\subsection{Y-Encoder and Query Module}
\label{appendix:yencoder-query}

The \texttt{Y-Encoder} $f_y$ is frozen E5-large-v2
\cite{wang2022e5} (\texttt{intfloat/e5-large-v2}, 335M parameters). It produces normalized embeddings for both complete
reference explanations and their constituent sentences:
\begin{align}
    \mathbf{s}_{y,i}
    &=
    \frac{
        f_y(\mathbf{y}_i)
    }{
        \|f_y(\mathbf{y}_i)\|_2
    },
    \\
    \mathbf{t}_{ij}
    &=
    \frac{
        f_y(y_{ij})
    }{
        \|f_y(y_{ij})\|_2
    }.
\end{align}
These embeddings are precomputed and remain fixed throughout training.
The complete reference embeddings supervise global semantic prediction,
while the sentence embeddings supervise query learning.
\texttt{Y-Encoder} is not loaded at inference.

In Stage~3b, four learned slot vectors cross-attend to a memory containing
the \texttt{Predictor}'s global hidden state and projected block-level
visual vectors. After residual feed-forward processing, independently
parameterized output projections map the slots into the semantic target
space. Normalization produces
\begin{equation}
    \mathbf{Q}_i
    =
    \left\{
        \mathbf{q}_{i1},
        \ldots,
        \mathbf{q}_{i4}
    \right\},
    \qquad
    \|\mathbf{q}_{ik}\|_2=1.
\end{equation}
The slots are exchangeable: no query index is assigned a predefined
forensic category, sentence position, or aspect label.

\subsection{Y-Decoder and Residual Adapter}
\label{appendix:ydecoder-adapter}

The \texttt{Y-Decoder} $d_{\psi}$ is
Qwen2.5-VL-3B-Instruct \cite{bai2025qwen25vl} (\texttt{Qwen/Qwen2.5-VL-3B-Instruct}). Its vision encoder and
pretrained language parameters remain frozen. Stage~3a trains a rank-$16$
language-side LoRA with scale $32$ on the attention and MLP projections.
The decoder receives the image through its native visual pathway and the
supplied verdict as the word \textit{real} or \textit{fake} in the prompt.

Stage~3c introduces query conditioning through a residual cross-attention
adapter. Let
$\mathbf{V}_i\in\mathbb{R}^{N_i\times d_{\mathrm{LM}}}$ denote the
decoder's existing visual-token embeddings and
$\mathbf{Q}_i\in\mathbb{R}^{4\times d_y}$ the learned semantic queries.
The adapter computes
\begin{equation}
\begin{split}
    \mathbf{V}'_i
    =
    \mathbf{V}_i
    +
    W_o\,
    \operatorname{MHA}
    \Big(
        W_v\operatorname{LN}_v(\mathbf{V}_i),
        W_q\operatorname{LN}_q(\mathbf{Q}_i),
        W_q\operatorname{LN}_q(\mathbf{Q}_i)
    \Big).
    \label{eq:residual_query_injection}
\end{split}
\end{equation}
Visual tokens act as attention queries, while the semantic vectors act as
keys and values. Only visual-token positions are modified, and the
original sequence length is preserved.

The output projection is initialized to zero,
\begin{equation}
    W_o=\mathbf{0},
    \qquad
    \mathbf{b}_o=\mathbf{0},
\end{equation}
so the adapter initially leaves the Stage~3a decoder unchanged. During
Stage~3c, only the residual adapter is trained; the decoder, its LoRA,
the predictor, and the query module remain locked.

At inference, the detector can return its verdict without loading
\texttt{Y-Encoder} or invoking \texttt{Y-Decoder}. If an explanation is
requested, the predicted verdict and query-conditioned visual tokens are
provided to the decoder for verbalization.

\subsection{Parameters}
\label{appendix:impl-parameters}

\Cref{tab:appendix-params} separates the two pathways. Detection depends on 28.35M trained parameters, the \texttt{X-Encoder} LoRA and the \texttt{Classifier}, all fixed by the end of Stage 1; the remaining 60.91M belong to the explanation pathway and are optimized only afterwards. Total trained parameters are 89.26M against roughly 3.65B of frozen released weights.

\begin{table}[!htbp]
\centering
\small
\caption{\textbf{Parameters of \ours.} Trained parameters by pathway and stage. }
\begin{tabular}{llrrl}
\toprule
\textbf{Module} & \textbf{Initialization} & \textbf{Total} & \textbf{Trained} & \textbf{Stage} \\
\midrule
\multicolumn{5}{l}{\textit{Detection pathway}} \\
\texttt{X-Encoder} & DINOv3 ViT-L/16 & 307M & -- & -- \\
\quad LoRA ($r{=}32$, $s{=}64$) & zero & 3.15M & 3.15M & 1 \\
\texttt{Classifier} & scratch & 25.20M & 25.20M & 1 \\
\midrule
\multicolumn{5}{l}{\textit{Explanation pathway}} \\
\texttt{Predictor} & scratch & 25.72M & 25.72M & 2, 3b \\
Query module & scratch & 3.68M & 3.68M & 3b \\
\texttt{Y-Decoder} & Qwen2.5-VL-3B-Instruct & 3.75B & -- & -- \\
\quad LoRA ($r{=}16$) & zero & 29.93M & 29.93M & 3a \\
Residual adapter & zero output & 1.58M & 1.58M & 3c \\
\midrule
\multicolumn{5}{l}{\textit{Training only, discarded at inference}} \\
\texttt{Y-Encoder} & E5-large-v2 & 335M & -- & -- \\
\bottomrule
\end{tabular}
\label{tab:appendix-params}
\end{table}

\subsection{Model Size Comparison}

FOX requires approximately $0.34$B parameters for detection and an additional $3.81$B for explanation ($4.15$B total), compared with approximately $7.06$B for FakeVLM, $7.71$B for SIDA-7B, and $8$B for LEGION. These figures represent complete inference models where available, or reported model sizes otherwise. FOX performs detection without loading the explanation pathway.
\section{Training Objectives and Hyperparameters}
\label{appendix:training}

This section gives the objectives and complete optimization settings for
the five sequential training steps described in
\cref{subsec:implementation-training}.

\subsection{Preliminaries}
\label{appendix:contrastive-objective}

Stages~2 and 3b use the same symmetric contrastive form. Given a similarity
matrix $\mathbf{A}\in\mathbb{R}^{B\times B}$ whose diagonal contains the
paired examples, define
\begin{equation}
    \mathcal{I}(\mathbf{A};\tau) = -\frac{1}{2B}\left[\sum_{i=1}^{B}\log\frac{\exp(A_{ii}/\tau)}{\sum_{m=1}^{B}\exp(A_{im}/\tau)} + \sum_{m=1}^{B}\log\frac{\exp(A_{mm}/\tau)}{\sum_{i=1}^{B}\exp(A_{im}/\tau)}\right].
    \label{eq:symmetric_infonce}
\end{equation}
The first term retrieves the paired text target for each image prediction;
the second retrieves the paired image prediction for each text target.

\subsection{Stage 1: Verdict Prediction}
\label{appendix:stage1}

The detector probability is
\begin{equation}
    p_{\phi}
    \left(
        \ell
        \mid
        \mathbf{x}_i
    \right)
    =
    \operatorname{softmax}
    \left(
        c_{\phi}
        \left(
            f_x(\mathbf{x}_i)
        \right)
    \right)_{\ell}.
\end{equation}
The \texttt{X-Encoder} LoRA and \texttt{Classifier} minimize
\begin{equation}
    \mathcal{L}_{1}
    =
    -\frac{1}{B}
    \sum_{i=1}^{B}
    \log
    p_{\phi}
    \left(
        \ell_i
        \mid
        \mathbf{x}_i
    \right).
    \label{eq:stage1_objective}
\end{equation}
No textual target participates in this stage.

\subsection{Stage 2: Joint Embedding Prediction}
\label{appendix:stage2}

For a minibatch of normalized predictions
$\hat{\mathbf{s}}_{y,i}$ and fixed reference embeddings
$\mathbf{s}_{y,i}$, define
\begin{equation}
    A^{\mathrm{global}}_{im}
    =
    \hat{\mathbf{s}}_{y,i}^{\top}
    \mathbf{s}_{y,m}.
\end{equation}
The \texttt{Predictor} is trained with
\begin{equation}
    \mathcal{L}_{2}
    =
    \mathcal{I}
    \left(
        \mathbf{A}^{\mathrm{global}};
        \tau_c
    \right).
    \label{eq:stage2_objective}
\end{equation}
The \texttt{X-Encoder}, \texttt{Classifier}, and
\texttt{Y-Encoder} remain fixed.

\subsection{Stage 3a: Decoder Adaptation}
\label{appendix:stage3a}

During decoder training, the ground-truth verdict $\ell_i$ is written into
the prompt. For the reference token sequence
$\mathbf{y}_i=(y_{i,1},\ldots,y_{i,T_i})$, the language-side LoRA
minimizes
\begin{equation}
    \mathcal{L}_{3a}
    =
    -\frac{1}{
        \sum_{i=1}^{B}T_i
    }
    \sum_{i=1}^{B}
    \sum_{r=1}^{T_i}
    \log
    p_{\psi}
    \left(
        y_{i,r}
        \mid
        y_{i,<r},
        \mathbf{x}_i,
        \ell_i
    \right).
    \label{eq:stage3a_objective}
\end{equation}
Only reference-explanation tokens contribute to the loss. This decoder is
unconditioned specifically with respect to the learned semantic queries;
it remains conditioned on the image and verdict-bearing prompt.

\subsection{Stage 3b: Relevance-Weighted Query Training}
\label{appendix:stage3b}

Stage~3b fine-tunes the \texttt{Predictor} together with the query module.
Let the reference explanation for image $i$ contain sentences
$\{y_{ij}\}_{j=1}^{J_i}$ with normalized targets
$\mathbf{t}_{ij}$.

\paragraph{Sentence relevance.}
Let $\mathcal{T}_{ij}$ contain the token positions belonging to sentence
$j$ in the complete reference. Its teacher-forced NLL under the frozen
Stage~3a decoder is
\begin{equation}
    n_{ij}
    =
    -\frac{1}{|\mathcal{T}_{ij}|}
    \sum_{r\in\mathcal{T}_{ij}}
    \log
    p_{\psi}
    \left(
        y_{i,r}
        \mid
        y_{i,<r},
        \mathbf{x}_i,
        \ell_i
    \right).
    \label{eq:sentence_nll}
\end{equation}
Let $F_{\mathrm{NLL}}$ be its empirical percentile over all training
sentences. We define
\begin{align}
    w^{\mathrm{raw}}_{ij}
    &=
    0.5+1.5F_{\mathrm{NLL}}(n_{ij}),
    \\
    \bar{w}_{ij}
    &=
    \frac{
        w^{\mathrm{raw}}_{ij}
    }{
        J_i^{-1}
        \sum_{r=1}^{J_i}
        w^{\mathrm{raw}}_{ir}
    }.
    \label{eq:relevance_weights}
\end{align}
The normalized weights have mean one within each explanation.

\paragraph{Set alignment.}
The compatibility between query set $\mathbf{Q}_i$ and sentence
$\mathbf{t}_{mj}$ is
\begin{equation}
    s
    \left(
        \mathbf{Q}_i,
        \mathbf{t}_{mj}
    \right)
    =
    \tau_s
    \log
    \sum_{k=1}^{K}
    \exp
    \left(
        \frac{
            \mathbf{q}_{ik}^{\top}
            \mathbf{t}_{mj}
        }{
            \tau_s
        }
    \right).
    \label{eq:query_sentence_score}
\end{equation}
The image-to-explanation score is
\begin{equation}
    S_{im}
    =
    \frac{
        \sum_{j=1}^{J_m}
        \bar{w}_{mj}
        s
        \left(
            \mathbf{Q}_i,
            \mathbf{t}_{mj}
        \right)
    }{
        \sum_{j=1}^{J_m}
        \bar{w}_{mj}
    },
    \label{eq:set_compatibility}
\end{equation}
and the set-alignment loss is
\begin{equation}
    \mathcal{L}_{\mathrm{set}}
    =
    \mathcal{I}
    \left(
        \mathbf{S};
        \tau_c
    \right).
    \label{eq:set_infonce}
\end{equation}

\paragraph{Collapse-resistant routing.}
The soft routing probability of sentence $j$ to query $k$ is
\begin{equation}
    a_{ikj}
    =
    \frac{
        \exp
        \left(
            \mathbf{q}_{ik}^{\top}
            \mathbf{t}_{ij}/\tau_s
        \right)
    }{
        \sum_{r=1}^{K}
        \exp
        \left(
            \mathbf{q}_{ir}^{\top}
            \mathbf{t}_{ij}/\tau_s
        \right)
    }.
    \label{eq:routing_probability}
\end{equation}
If the queries are identical, $a_{ikj}=1/K$ and every slot receives the
same sentence-level signal. We therefore anneal $\tau_s$ from a soft
initial value to a sharper final value and add three complementary terms.

Sentence-level ownership is encouraged by
\begin{equation}
    \mathcal{L}_{\mathrm{own}}
    =
    -\frac{1}{B}
    \sum_{i=1}^{B}
    \frac{
        \sum_j
        \bar{w}_{ij}
        \log(\max_k a_{ikj})
    }{
        \sum_j\bar{w}_{ij}
    }.
    \label{eq:ownership_loss}
\end{equation}
To discourage a single query from owning every sentence, define
\begin{align}
    h_{ij}
    &=
    \arg\max_k a_{ikj},
    \\
    f_{ik}
    &=
    \frac{
        \sum_j
        \bar{w}_{ij}\mathbb{I}[h_{ij}=k]
    }{
        \sum_j\bar{w}_{ij}
    },
    \\
    P_{ik}
    &=
    \frac{
        \sum_j
        \bar{w}_{ij}a_{ikj}
    }{
        \sum_j\bar{w}_{ij}
    },
\end{align}
and
\begin{equation}
    \mathcal{L}_{\mathrm{bal}}
    =
    \frac{1}{B}
    \sum_{i=1}^{B}
    K
    \sum_{k=1}^{K}
    f_{ik}P_{ik}.
    \label{eq:balance_loss}
\end{equation}
The hard assignments in $f_{ik}$ are treated as constants during
backpropagation.

Finally, normalized queries are separated geometrically using
\begin{equation}
    \mathcal{L}_{\mathrm{orth}}
    =
    \frac{1}{
        BK(K-1)
    }
    \sum_{i=1}^{B}
    \sum_{\substack{k,r=1\\k\neq r}}^{K}
    \left(
        \mathbf{q}_{ik}^{\top}
        \mathbf{q}_{ir}
    \right)^2.
    \label{eq:orthogonality_loss}
\end{equation}
Independently parameterized slot-output projections provide an additional
architectural symmetry break.

\paragraph{Global retention and complete objective.}
Let $\hat{\mathbf{s}}^{\mathrm{FT}}_{y,i}$ be the global output of the
fine-tuned predictor and define
\begin{equation}
    G_{im}
    =
    \left(
        \hat{\mathbf{s}}^{\mathrm{FT}}_{y,i}
    \right)^{\top}
    \mathbf{s}_{y,m}.
\end{equation}
The global retention term is
\begin{equation}
    \mathcal{L}_{\mathrm{global}}
    =
    \mathcal{I}
    \left(
        \mathbf{G};
        \tau_c
    \right).
\end{equation}
The complete Stage~3b objective is
\begin{equation}
    \mathcal{L}_{3b}
    =
    \mathcal{L}_{\mathrm{set}}
    +
    \lambda_{\mathrm{global}}
    \mathcal{L}_{\mathrm{global}}
    +
    \lambda_{\mathrm{own}}
    \mathcal{L}_{\mathrm{own}}
    +
    \lambda_{\mathrm{bal}}
    \mathcal{L}_{\mathrm{bal}}
    +
    \lambda_{\mathrm{orth}}
    \mathcal{L}_{\mathrm{orth}}.
    \label{eq:stage3b_objective}
\end{equation}

\subsection{Stage 3c: Residual-Adapter Training}
\label{appendix:stage3c}

With the predictor, query module, and Stage~3a decoder locked, only the
residual adapter parameters $\omega$ are trained:
\begin{equation}
    \mathcal{L}_{3c}
    =
    -\frac{1}{
        \sum_{i=1}^{B}T_i
    }
    \sum_{i=1}^{B}
    \sum_{r=1}^{T_i}
    \log
    p_{\psi,\omega}
    \left(
        y_{i,r}
        \mid
        y_{i,<r},
        \mathbf{x}_i,
        \ell_i,
        \mathbf{Q}_i
    \right).
    \label{eq:stage3c_objective}
\end{equation}
Only reference tokens contribute to this loss. The residual adapter learns
how the frozen decoder should consume $\mathbf{Q}_i$ but cannot modify the
queries, decoder, or detector.

\paragraph{Verdict sources.}
During decoder adaptation, sentence-NLL computation, and residual-adapter
training, the prompt contains the ground-truth verdict:
\begin{equation}
    \tilde{\ell}^{\mathrm{train}}_i
    =
    \ell_i.
\end{equation}
At validation and inference, it contains the locked detector prediction:
\begin{equation}
    \tilde{\ell}^{\mathrm{eval}}_i
    =
    \hat{\ell}_i
    =
    \mathbb{I}
    \left[
        \operatorname{softmax}
        \left(
            c_{\phi}(f_x(\mathbf{x}_i))
        \right)_1
        \geq\gamma
    \right],
\end{equation}
where $\gamma$ is the operating threshold of the corresponding evaluation
protocol. The explanation pathway consumes this verdict as a fixed word
and cannot propagate gradient into the detector.

\subsection{Optimization}
\label{appendix:optimization-settings}

All experiments use seed $42$. The five optimization steps are summarized
below.

\begin{table*}[!htbp]
\centering
\small
\caption{\textbf{Optimization settings for the five sequential training steps.}
Products in the effective-batch column denote physical batch size times
gradient accumulation.}
\setlength{\tabcolsep}{5pt}
\begin{tabular}{lccccc}
\toprule
\textbf{Stage}
& \textbf{Optimizer}
& \textbf{Learning rate}
& \textbf{Weight decay}
& \textbf{Effective batch}
& \textbf{Duration} \\
\midrule
1
& AdamW
& $1\times10^{-4}$
& $1\times10^{-2}$
& $16\times4=64$
& 1 epoch \\
2
& SAM--AdamW
& $3\times10^{-4}$
& $5\times10^{-2}$
& $2048$
& 24 epochs \\
3a
& AdamW
& $1\times10^{-4}$
& $1\times10^{-2}$
& $4\times8=32$
& 1 epoch \\
3b
& AdamW
& $3\times10^{-4}$
& $1\times10^{-2}$
& $128$
& 15 epochs \\
3c
& AdamW
& $1\times10^{-4}$
& $1\times10^{-2}$
& $4\times8=32$
& 1 epoch \\
\bottomrule
\end{tabular}
\label{tab:complete_optimization}
\end{table*}

Stage~1 uses class-balanced undersampling, a $512$-step linear warm-up,
and checkpoint evaluation at fixed intervals on the external Chameleon
validation set; the checkpoint with maximum validation AUC is selected.
Stage~2 uses SAM radius $\rho=0.05$, a $500$-step warm-up, and cosine
decay. Stages~3a and 3c use class-balanced undersampling and cosine decay.
Stage~3b uses class-balanced cycling and cosine decay.

For Stage~3b, we use
\begin{equation}
    K=4,
    \qquad
    \tau_c=0.05,
    \qquad
    \tau_s:0.5\rightarrow0.05,
\end{equation}
with routing temperature annealed over the first seven epochs and held
fixed thereafter. The objective coefficients are
\begin{equation}
    \lambda_{\mathrm{global}}=0.25,
    \qquad
    \lambda_{\mathrm{own}}=0.16,
    \qquad
    \lambda_{\mathrm{bal}}=0.14,
    \qquad
    \lambda_{\mathrm{orth}}=0.15.
\end{equation}
The final query configuration uses slot-specific output projections and
no slot dropout.

The decoder LoRA uses rank $16$, scale $32$, and dropout $0.05$. Decoder
targets are capped at $360$ tokens. The residual adapter uses attention
width $256$ with eight heads and adds $1.58$M trainable parameters.
Explanation generation is greedy and produces at most $180$ new tokens.

Across its disjoint training stages, \ours optimizes $89.26$M parameters,
of which $28.35$M define the verdict pathway. Approximately $3.65$B
released backbone parameters remain frozen.

\section{Data and Evaluation Protocol}
\label{appendix:evaluation}

\subsection{XPlainVerse Training Dataset}

\ours and the \mbox{XPV} baselines are trained using XPlainVerse~\cite{narang2026xplainverse}. The publicly available release provides training and validation splits. The training split described in \cref{tab:xpv-training-composition} contains 360,000 examples from three generation sources. The distribution is highly concentrated in Gemini-2.0-Flash. None of the exact generator versions in this training inventory appears as an evaluation generator. The public XPV validation split contains 110,000 images. We randomly sample 10,000 of these images and use this fixed subset as the XPV test set throughout the study; the XPV results reported in \Cref{tab:explanation} are computed on this subset. Explanation-side ablations instead use a 20,000-image holdout from the training split, preventing leakage into the final XPV test set. This separation is important because the public validation split exhibits a distribution shift from training~\cite{narang2026xplainverse} and is therefore reserved for final evaluation only.

\begin{table}[!htbp]
    \centering
    \small
    \caption{\textbf{Generator composition of the synthetic XPV training
    data.} Percentages are relative to the 360000 synthetic examples.}
    \label{tab:xpv-training-composition}
    \setlength{\tabcolsep}{7pt}
    \begin{tabular}{@{}lrr@{}}
        \toprule
        \textbf{Generation source} & \textbf{Examples} & \textbf{Share} \\
        \midrule
        Gemini-2.0-Flash & 354,256 & 98.40\% \\
        ICEdit           &   3050 &  0.85\% \\
        GPT-Image-1      &   2694 &  0.75\% \\
        \midrule
        Total            & 360,000 & 100.00\% \\
        \bottomrule
    \end{tabular}
\end{table}

The \mbox{XPV} versions of SPAI, RINE, B-Free and DINO-MAC use the same XPV
training data and the same external Chameleon validation protocol as \ours.
Consequently, their comparison with \ours controls for access to the XPV
training distribution and for the checkpoint-selection data.

\subsection{Evaluation Datasets}

The detection benchmark combines data from the SPAI
any-resolution protocol~\cite{karageorgiou2025spai} and
OpenFake~\cite{livernoche2025openfake}. The main paper comparison contains
14 generators: 10 from SPAI whose outputs exceed 0.5 megapixels and four
held-out OpenFake generators. The expanded appendix evaluation adds 19
generators, producing a total of 33 fake image sources. The appendix-only
generators do not contribute to the mean reported in the main paper detection
table.

Both evaluations use the same five real image sources. RAISE and FODB come
from the SPAI protocol, while COCO, ImageNet, and OpenImages provide the
shifted real image distributions introduced through OpenFake. Each real
source contains 1000 images.

\begin{table*}[!htbp]
    \centering
    \small
    \caption{\textbf{Detection evaluation inventory.} Parentheses give the number of evaluated images. Rows marked \emph{SPAI protocol} are not drawn from a single released benchmark: we follow the collection protocol of~\cite{karageorgiou2025spai} and gather images from its specified sources. OpenFake~\cite{livernoche2025openfake} is a released benchmark. The main paper reports 14 generators; the appendix adds 19. The same 5 real sources are used in both evaluations.}
    \label{tab:detection-evaluation-inventory}
    \setlength{\tabcolsep}{4pt}
    \renewcommand{\arraystretch}{1.15}
    \begin{tabular}{@{}p{0.10\linewidth}p{0.14\linewidth}
                    p{0.61\linewidth}r@{}}
        \toprule
        \textbf{Reported} & \textbf{Protocol / source} & \textbf{Generators or real-image sources}
        & \textbf{Total images}\\
        \midrule
        Main
        & SPAI protocol~\cite{karageorgiou2025spai}
        & From Synthbuster~\cite{bammey2023synthbuster}: SD2 (1000), SDXL (1000),
          MJv5 (1000), DALLE2 (1000), DALLE3 (1000), and Firefly (1000).
          From~\cite{latentspacer2024flux}: Flux (1000).
          From~\cite{saq1b2024midjourney}: MJv6.1 (638).
          From~\cite{meisinger2024laion}: SD3 (1000).
          From~\cite{kang2023gigagan}: GigaGAN (1000)
        & 9638 \\
        Main
        & OpenFake~\cite{livernoche2025openfake}
        & MJ7 (1000), GPT-I2 (474), GPT-I1.5 (1000),
          and NB-Pro (386)
        & 2860 \\
        \midrule
        Appendix
        & SPAI protocol~\cite{karageorgiou2025spai}
        & From Synthbuster~\cite{bammey2023synthbuster}: GLIDE (1000),
          SD1.3 (1000), and SD1.4 (1000)
        & 3000 \\
        Appendix
        & OpenFake~\cite{livernoche2025openfake}
        & Aurora-20-1-25 (282), ERNIE-Image (315),
          ERNIE-Image-Turbo (687), FLUX.2-Klein-9B (1000),
          Frames-23-1-25 (250), Halfmoon-4-4-25 (190),
          Ideogram-2.0 (282), Illustrious (1000),
          Lumina-17-2-25 (543), Recraft-v2 (282),
          Recraft-v3 (1000), Seedream-v5.0 (372),
          Sora-2 (557), Veo-3 (1000),
          Wan-Video-2.5 (1000), and Z-Image-Turbo (1000)
        & 9760 \\
        \midrule
        Both
        & SPAI protocol~\cite{karageorgiou2025spai}
        & RAISE~\cite{nguyean2015raise} (1000), FODB~\cite{hadwiger2020fodb} (1000),
          ImageNet~\cite{deng2009imagenet} (1000), COCO~\cite{lin2014coco} (1000),
          and OpenImages~\cite{kuznetsova2020openimages} (1000)
        & 5000 \\
        \bottomrule
    \end{tabular}
\end{table*}

The main paper benchmark therefore contains 12498 fake images from 14
generators and 5000 real images. The expanded evaluation contains 25,258
fake images from 33 generators and the same 5000 real images, for 30,258
unique evaluation images in total. Generator datasets smaller than 1000
images use all available valid images; the remaining datasets are capped at
1000 images.

\subsection{Detection Score Aggregation}

We evaluate each fake generator against each real image source separately.
For a particular generator, a detector first assigns one fake-likelihood
score to every image from that generator and to every image from each of
COCO, ImageNet, OpenImages, RAISE, and FODB. We then construct five binary
comparisons: the generator against COCO, against ImageNet, against
OpenImages, against RAISE, and against FODB. In every comparison, synthetic
images are the positive class.

An AUC is calculated independently for each of these five comparisons. The
score shown for that generator is the unweighted mean of the five resulting
AUC values. We deliberately do not pool the five real datasets into a single
negative set: separate evaluation prevents the composition of a larger real
pool from obscuring sensitivity to a particular real-image distribution.

The final \emph{Mean} column is also unweighted across generators. In the
main paper, it averages the 14 displayed generator scores; in the expanded
appendix table, it averages all 33 generator scores. Thus, every real source
has equal influence within a generator score, and every generator has equal
influence on the final mean, irrespective of its number of available images.

The robustness evaluation follows the same aggregation process independently
at every perturbation setting. A given perturbation is applied to both the
fake and real images before scoring, and the pairwise AUC always compares
images subjected to the exact same perturbation and severity.

\subsection{Text Similarity Metrics}
Each generated explanation is compared with its single reference explanation, and the dataset-level score is the arithmetic mean of the per-example scores.

\textbf{ROUGE-L.}
Both texts are Unicode-case-folded and tokenized into words and individual punctuation symbols using the regular expression \texttt{\textbackslash w+|[\^{}\textbackslash w\textbackslash s]}. No stemming is applied. Let \(m\) and \(n\) be the numbers of prediction and reference tokens, respectively, and let \(L\) be the length of their longest common subsequence. We compute
\[
P_{\mathrm{LCS}}=\frac{L}{m},\qquad
R_{\mathrm{LCS}}=\frac{L}{n},\qquad
\mathrm{ROUGE\text{-}L}_{F_1}
=\frac{2P_{\mathrm{LCS}}R_{\mathrm{LCS}}}
       {P_{\mathrm{LCS}}+R_{\mathrm{LCS}}}.
\]
The score is set to zero if either text is empty.

\textbf{Cosine semantic similarity (CSS).}
The raw prediction and reference texts, without task-specific prefixes, are encoded independently using
\texttt{sentence-transformers/paraphrase-MiniLM-L6-v2}, following~\cite{kang2025legionlearninggroundexplain}, with the model's standard sentence pooling and a maximum sequence length of 512 tokens. If \(\mathbf{z}_p\) and \(\mathbf{z}_r\) denote their sentence embeddings, CSS is
\[
\mathrm{CSS}(p,r)=
\frac{\mathbf{z}_p^{\top}\mathbf{z}_r}
     {\lVert\mathbf{z}_p\rVert_2\lVert\mathbf{z}_r\rVert_2}.
\]

\textbf{BERTScore.}
We use the English \texttt{microsoft/deberta-xlarge-mnli} encoder (layer 40) with a maximum sequence length of 512 tokens. Baseline rescaling and inverse-document-frequency weighting are disabled. Let \(\mathbf{h}^{p}_i\) and \(\mathbf{h}^{r}_j\) be the contextual embeddings of the non-special subword tokens in the prediction and reference, and let
\[
s_{ij}=
\frac{(\mathbf{h}^{p}_i)^\top\mathbf{h}^{r}_j}
     {\lVert\mathbf{h}^{p}_i\rVert_2\lVert\mathbf{h}^{r}_j\rVert_2}.
\]
BERTScore precision and recall use greedy maximum-similarity matching:
\[
P_{\mathrm{BERT}}=\frac{1}{m}\sum_{i=1}^{m}\max_j s_{ij},
\qquad
R_{\mathrm{BERT}}=\frac{1}{n}\sum_{j=1}^{n}\max_i s_{ij},
\]
and
\[
\mathrm{BERTScore}\text{-}F_1=
\frac{2P_{\mathrm{BERT}}R_{\mathrm{BERT}}}
     {P_{\mathrm{BERT}}+R_{\mathrm{BERT}}}.
\]
Raw scores are retained for aggregation; where percentage-scale results are shown, their dataset means are multiplied by \(100\).

\section{Extended Results}
\label{appendix:results}

\subsection{Detection}
\label{appendix:results-detection}

Recent DINOv3-based systems have achieved top rankings in  deepfake detection 
competitions~\cite{qu2026dinomac,narang2026explainabledeepfakedetectionchallenge}. However, the trained checkpoints or their exact training data required to evaluate these systems under our common protocol are not publicly available, so they are not included in the quantitative comparison. \Cref{tab:detection-auc,tab:detection-acc,tab:detection-ap} extends the evaluation to all 33 unseen generators. \ours remains strongest overall, achieving mean AUC, AP, and fixed threshold accuracy of $96.1$, $95.0$, and $89.8$, respectively. The broader suite nevertheless exposes generator-specific limitations. Most notably, performance on Flux2-K9 falls to $53.6$ AUC, $61.3$ AP, and $53.8$ accuracy. Firefly presents a different failure mode: its $90.0$ AUC and $90.4$ AP fall to $58.7$ accuracy at the fixed threshold of $0.5$, indicating poor score calibration instead of a complete loss of ranking ability. Thus, the aggregate gains do not imply uniform transfer, and calibration under generator shift remains an open problem.

\subsection{Generalization to newer generators}
\label{appendix:cham-generalization}
Chameleon was introduced in June 2024 and publicly released in December 2024, with no documented subsequent expansion. To assess generalization beyond potential generator overlap, we evaluate subsets released after these dates. The post-December subset comprises 14 generators: Midjourney v7, GPT-Image-2, GPT-Image-1.5, FLUX.2 Klein, Sora 2, Veo 3, Wan 2.5, Halfmoon, ERNIE-Image, ERNIE-Image-Turbo, Lumina-Image 2.0, Nano Banana Pro, Seedream 5.0, and Z-Image Turbo. The post-June subset additionally includes Midjourney v6.1, FLUX.1, Ideogram 2.0, Illustrious, Recraft v3, Frames, and Aurora, totaling 21 generators. As shown in \cref{tab:temporal-generalization}, \ours achieves 94.8 and 93.4 mean AUC on the respective subsets, outperforming the strongest XPV-retrained baseline by 4.3 and 4.7 points.

\begin{table}[!htbp]
\centering
\small
\caption{\textbf{Temporal generalization.}
Mean AUC ($\times100$) across 33 generators
and subsets released after Chameleon's
publication (June 2024) and dataset release
(December 2024). All methods share the
XPV training and Chameleon checkpoint-selection
protocol.}
\label{tab:temporal-generalization}
\setlength{\tabcolsep}{5pt}
\begin{tabular}{lccc}
\toprule
\textbf{Method}
& \textbf{All 33}
& \textbf{Post-Jun.\ (21)}
& \textbf{Post-Dec.\ (14)} \\
\midrule
SPAI$_{\mathrm{XPV}}$
& 82.2 & 82.7 & 82.5 \\
RINE$_{\mathrm{XPV}}$
& 85.6 & 86.0 & 83.7 \\
B-Free$_{\mathrm{XPV}}$
& \underline{91.6} & \underline{90.5}
& \underline{88.8} \\
DINO-MAC$_{\mathrm{XPV}}$
& 91.5 & 88.0 & 85.6 \\
\midrule
\textbf{\ours}
& \textbf{96.1} & \textbf{94.8}
& \textbf{93.4} \\
\bottomrule
\end{tabular}
\end{table}



\subsection{Qualitative Examples and Failure Cases}
\label{appendix:results-qualitative}

We complement the main quantitative results with qualitative examples probing \ours\ along three axes: generalization to unseen generators, explanation quality across the BERTScore distribution, and characteristic failure modes.

\textbf{Generalization to unseen generators.} Fig.~\ref{fig:appendix_openfake} show two OpenFake examples for each of five recent image/video generators (GPT-Image-1.5, GPT-Image-2, Midjourney-7, Veo-3, and Nano Banana Pro), none of which appear in our training data. \ours\ correctly flags all 10 images as fake, including two examples from GPT-Image-2, the single hardest generator in this set. For each example we report the \ours score together with the model's predicted-verdict explanation. This indicates that \ours's forensic reasoning transfers to novel synthesis artifacts (e.g., fused fingers on a sword grip, an airbrushed wolf pelt, an anatomically implausible duck neck, flattened compositing on a man surrounded by cats, or unnaturally smooth gingerbread icing) without any generator-specific fine-tuning. Because OpenFake has no reference explanations, no ground-truth column is reported for this figure.

\textbf{Explanation quality across the BERTScore distribution.} Fig.~\ref{fig:appendix_pure_success} shows four correctly classified XPV examples, spanning both fake and real ground truth and both high- and low-BERTScore explanations. In every case the verdict is correct, and even the low-BERTScore explanations (BERTScore F1 as low as 0.65) identify a genuine, independently verifiable cue (e.g., the floating figure and mismatched bicycle contact shadows in the red jacket example, or the wrinkled plastic bag in the bicycle basket example) that simply does not lexically overlap with the reference. This supports our claim in the main paper that BERTScore underestimates semantically correct but differently-phrased explanations.

\textbf{Failure modes.} Fig.~\ref{fig:appendix_pure_failure} shows four representative errors, two of each error type. In the fake labeled birthday cake and freight car examples, \ours\ predicts ``real'' by anchoring on a plausible but ultimately incorrect cue (bumpy frosting, muddy wheels) instead of the actually manipulated region that was flagged (the flat, two-dimensional ``ALEX'' text overlay; the pasted-on newspaper texture wrapped around the curved tank car end). In the real labeled cyclist and chef examples, \ours\ predicts ``fake'' by misreading ordinary capture artifacts, motion blur and a grainy, low-light exposure, as signs of oversmoothed, synthetic skin texture. Taken together, these failure cases suggest that \ours's remaining errors are concentrated at the boundary between generative artifacts and conventional photographic artifacts (blur, grain, compression, print/graphic overlays), and are not random or uninterpretable. We see explicitly disentangling these two artifact families as promising future work.

\section{Extended Ablation Study}
\label{appendix:ablations}

\subsection{Implementation Details of \texorpdfstring{\Cref{tab:backbone}}{Table 3}}
\label{appendix:ablation-backbone}

All eight backbones in \Cref{tab:backbone} use public pretrained checkpoints and remain fully frozen. We read the final four transformer blocks and construct one representation per block from the token types exposed by each checkpoint, concatenating the patch token mean, \texttt{[cls]} token, and register tokens where available, or otherwise using the available subset down to the patch token mean alone. The resulting per-block width therefore ranges from $1024$ dimensions to $6144$ for DINOv2 and DINOv3; each representation is independently projected to the common trunk width $d_h=512$, so every backbone supplies the same four-token sequence to the downstream transformer. The parameter counts in \Cref{tab:appendix-trunk} correspond to the $6144$-dimensional DINOv3 readout and are thus the largest, since only the input projections depend on backbone-specific width. Seven backbones contain approximately $300$--$400$M parameters, with I-JEPA as the scale exception at approximately $600$M (ViT-H). Thus, the comparison controls the downstream readout, training data, and evaluation protocol, but not the pretrained corpus or exact backbone scale, which are properties of the released checkpoints.

\begin{itemize}[nosep, leftmargin=*]
    \item \textbf{CLIP}~\cite{radford2021learning}: \texttt{vit\_large\_patch14\_clip\_quickgelu\_336.openai}, ViT-L/14 pretrained by OpenAI with vision-language contrastive supervision, input resolution 336. Readout is patch mean concatenated with the \texttt{[cls]} token, taken from blocks 20-23 of 24.
    \item \textbf{SigLIP}~\cite{zhai2023sigmoid}: \texttt{vit\_so400m\_patch14\_siglip\_384.webli}, a shape-optimized ViT-So400m/14 pretrained with vision-language sigmoid contrastive loss on WebLI, input resolution 384. This checkpoint exposes no prefix token through our interface, so readout is the patch mean alone, taken from the last four of its 27 blocks.
    \item \textbf{AIMv2}~\cite{fini2025aimv2}: \texttt{aimv2\_large\_patch14\_336.apple\_pt}, ViT-L/14 pretrained by Apple with a multimodal autoregressive objective, input resolution 336. No prefix token is exposed, so readout is the patch mean alone, taken from blocks 20-23 of 24.
    \item \textbf{MAE}~\cite{he2022mae}: \texttt{vit\_large\_patch16\_224.mae}, ViT-L/16 pretrained as a masked autoencoder, input resolution 224. Readout is patch mean concatenated with the \texttt{[cls]} token, taken from blocks 20-23 of 24.
    \item \textbf{I-JEPA}~\cite{assran2023ijepa}: \texttt{vit\_huge\_patch14\_gap\_224.in1k\_ijepa}, ViT-H/14 pretrained on ImageNet-1k with the image JEPA objective, input resolution 224. This is our only ViT-Huge backbone (\textasciitilde600M parameters). No prefix token is exposed, so readout is the patch mean alone, taken from blocks 28-31 of its 32 blocks.
    \item \textbf{V-JEPA 2}~\cite{assran2025vjepa2}: \texttt{facebook/vjepa2-vitl-fpc64-256}, a video ViT-L pretrained with the video JEPA objective, input resolution 256. Since it is a video model, we repeat the static image across 16 frames and mean-pool over all resulting space-time tokens, taken from blocks 20-23 of 24.
    \item \textbf{DINOv2}~\cite{oquab2023dinov2}: \texttt{vit\_large\_patch14\_reg4\_dinov2.lvd142m}, ViT-L/14 with four register tokens, pretrained on LVD-142M with self-distillation, input resolution 392. Readout is patch mean concatenated with the \texttt{[cls]} token and the four register tokens, taken from blocks 20-23 of 24.
    \item \textbf{DINOv3}~\cite{simeoni2025dinov3}: \texttt{vit\_large\_patch16\_dinov3.lvd1689m}, ViT-L/16 with four register tokens, pretrained on LVD-1689M with self-supervised learning, input resolution 384. Readout is patch mean concatenated with the \texttt{[cls]} token and the four register tokens, taken from blocks 20-23 of 24.
\end{itemize}

\subsection{Readout Capacity}
\label{appendix:ablation-trunk}

\emph{Do the backbone results depend on downstream readout capacity?}
\Cref{tab:appendix-trunk} varies only the classifier readout while
holding the frozen DINOv3 representation fixed. All readouts perform
similarly on the five unseen GenImage generators, but the linear head
transfers markedly worse to Chameleon. Transformer depth then quickly
saturates: the four- and eight-layer variants perform nearly identically,
while the four-layer trunk uses substantially fewer parameters. We
therefore use the four-layer transformer throughout the backbone
comparison.

\begin{table}[!htbp]
\centering
\small
\caption{\textbf{How much readout capacity does the comparison need?} AUC with a frozen DINOv3 ViT-L/16 and only the readout varied, trained on XPlainVerse and tested on unseen generators. \textbf{GenImage} averages five unseen GenImage generators (SD1.5, BigGAN, VQDM, MJv4, Wukong); Chameleon is reported separately. Parameter counts correspond to the $6144$-dimensional DINOv3 readout.}
\label{tab:appendix-trunk}
\setlength{\tabcolsep}{3pt}
\begin{tabular}{lrcc}
\toprule
\textbf{Readout} & \textbf{Params} & \textbf{GenImage} & \textbf{Cham} \\
\midrule
Linear & 0.05M & 98.6 & 68.9 \\
Transformer-2L & 18.90M & \cellcolor{gray!10}\textbf{98.9} & 76.3 \\
Transformer-4L & 25.20M & 98.8 & \textbf{79.9} \\
Transformer-8L & 37.81M & 98.7 & 79.6 \\
\bottomrule
\end{tabular}
\end{table}

\subsection{Implementation Details of \texorpdfstring{\Cref{tab:ablation_decoupling}}{Table 5}}
\label{appendix:ablation-decoupling}

\noindent\textbf{Shared trunk.}
Verdict prediction and joint-embedding prediction share one visual trunk and
are optimized jointly with $\mathcal{L}_{1}+\mathcal{L}_{2}$.
In \textit{Shared trunk, early stop}, checkpoint selection follows detection
performance on Chameleon. In \textit{Shared trunk, late stop}, training instead
continues for the full InfoNCE schedule, allowing the embedding objective to
converge further. The former favors detection performance, whereas the latter
favors explanation quality.

\noindent\textbf{Single end-to-end pipeline.}
This variant combines shared-trunk verdict and embedding learning with
concurrent Stage~3a decoder adaptation in one training schedule, optimizing
$\mathcal{L}_{1}+\mathcal{L}_{2}+\mathcal{L}_{3a}$. Stage~3b query training and
Stage~3c residual adapter training are subsequently applied unchanged.

\subsection{Implementation Details of \texorpdfstring{\Cref{tab:ablation_explanation}}{Table 7}}
\label{appendix:ablation-explanation}

We first compare three VLMs using Stage~3a only: each model is fine-tuned to generate an explanation from the image and verdict, without semantic conditioning. Qwen performs best and is used thereafter. In Stage~3a+3c, the Stage~2 Predictor's global explanation embedding is passed directly to the residual adapter, without the Stage~3b query module. The full model instead conditions the adapter on four sentence-level queries learned in Stage~3b. Its clear improvement shows that structured query conditioning is more effective than either the unconditioned decoder or a single global semantic representation.

\subsection{Implementation Details of \texorpdfstring{\Cref{tab:ablation_queries}}{Table 8}}
\label{appendix:ablation-queries}

For the shuffled-query control in \Cref{tab:ablation_queries}, we replace
each image's complete query set with that of a different image from the
same authenticity class, while keeping the image and verdict prompt
unchanged. We shuffle whole query sets rather than individual slots.
This preserves class information, query count, and the class-conditional
query distribution while breaking only the correspondence between the
image and its forensic guidance. The resulting drop in
BERTScore therefore reflects the loss of image-specific evidence rather
than a mismatch between real and fake queries.

\subsection{Semantic Pretraining}
\label{appendix:ablation-stage2-pretraining}

\emph{Does query learning benefit from semantic pretraining of the Predictor?}
We compare the Stage~2-pretrained Predictor with an identically
parameterized Predictor initialized randomly before Stage~3b. Both
variants subsequently undergo the same Stage~3b query training and
Stage~3c adapter training. \Cref{tab:ablation_stage2_pretraining} shows
that removing Stage~2 substantially reduces explanation quality,
indicating that the semantic representation learned before query
training provides an important initialization. We therefore retain
Stage~2 pretraining.

\begin{table}[!htbp]
\centering
\small
\caption{\textbf{Effect of semantic predictor pretraining.} The two variants differ only in the initialization used at the beginning of Stage 3b.}
\label{tab:ablation_stage2_pretraining}
\setlength{\tabcolsep}{3pt}
\begin{tabular}{lc}
\toprule
\textbf{Predictor initialization}
& \textbf{BERTSc.} \\
\midrule
Random init. (no Stage 2)
& 68.3 \\
Stage 2 semantic pretraining (ours)
& \cellcolor{gray!10}\textbf{72.1} \\
\bottomrule
\end{tabular}
\end{table}

\subsection{Predictor Objective}
\label{appendix:ablation-predictor-objective}

\emph{How should detector evidence be predicted in the semantic embedding space?} Direct regression provides a simple alignment objective, whereas contrastive learning preserves relative structure using negatives~\cite{oord2018cpc, gao2021simcse}. \Cref{tab:ablation_objective} shows that InfoNCE provides the stronger explanation representation, consistent with the anisotropic geometry of pretrained text embeddings~\cite{ethayarajh2019contextual, gao2019representation}. We therefore use contrastive embedding prediction in Stage~2.

\begin{table}[!htbp]
\centering
\small
\caption{\textbf{Ablation on the predictor objective.} Direct cosine regression is compared with contrastive embedding prediction.}
\label{tab:ablation_objective}
\setlength{\tabcolsep}{3pt}
\begin{tabular}{lc}
\toprule
\textbf{Stage 2: Objective}
& \textbf{BERTSc.} \\
\midrule
Cosine regression
& 65.6 \\
InfoNCE (ours)
& \cellcolor{gray!10}\textbf{72.1} \\
\bottomrule
\end{tabular}
\end{table}

\section{Discussion}
\label{appendix:discussion}

\subsection{Scope of the Claims}
\label{appendix:discussion-scope}

Our detection results are an investigation of what generalizes without claiming superiority for pure-vision pretraining. 
Under a matched readout, DINOv3's view-consistency representation transfers best, while the weaker MAE, I-JEPA, and V-JEPA 2 results show that removing language supervision alone is insufficient; the discriminating property is view consistency, and not simply the absence of a text objective. The adaptation recipe follows the parameter-efficient configuration that has proven effective in recent robust detection challenges~\cite{qu2026dinomac,hopf2026robustdeepfakedetectionntire}, and we adopt it deliberately: our aim is to identify which choices along this pathway carry cross-generator transfer, evaluated under a single controlled protocol. Within that pathway, checkpoint selection is the choice that has remained unspecified. Table~\ref{tab:adaptation} illustrates this: XPV validation AUC is nearly saturated across adaptation strategies, despite substantial differences in cross-generator transfer, making internal validation unreliable for checkpoint selection. Selecting the same LoRA trajectory on an independent external validation distribution recovers the gap, and makes explicit a criterion that competition entries approximate through oracle or unreported selection. We report this as a design requirement with a stated cost, namely a labeled external validation distribution. Finally, locking the detector before explanation training guarantees that explanation gradients cannot alter its verdict without claiming that  this architectural guarantee is in any way implying a causal link between the generated text and the classifier's computation.

\subsection{Limitations}
\label{appendix:discussion-limitations}

An embedding is not an explanation. Because \texttt{Y-Decoder} receives the image and locked verdict, the predicted queries should be interpreted as forensic guidance for post-hoc verbalization, but not as a complete account of the decision. Explanation quality is evaluated against automatically constructed references and through similarity metrics, which measure agreement with a reference, but this does not imply factual correctness or causal faithfulness. Although LOKI evaluates transfer beyond XPV, human evaluation remains necessary. \ours also produces a whole-image verdict and does not localize manipulated regions. Its detector is trained on locally edited images but evaluated largely on fully synthetic generators, leaving possible provenance and dataset biases despite strong cross-generator transfer. Finally, the four unordered queries have no predefined semantic roles, so their aggregate contribution is measurable but individual queries are not directly interpretable.

\section{Reproducibility}
\label{appendix:reproducibility}

To ensure full reproducibility, we provide comprehensive architectural details (\Cref{appendix:components}), implementation details (\Cref{appendix:training}), and evaluation protocols (\Cref{appendix:evaluation}). Every component of \ours is initialized from a publicly available checkpoint, and all evaluations use public datasets under their published protocols~\cite{karageorgiou2025spai, livernoche2025openfake, zhu2023genimage, yan2024sanity}. Fixed random seeds are used throughout, though minor variations may occur due to non-deterministic CUDA operations.

\clearpage
\begin{table}[t]
\centering
\small
\setlength{\tabcolsep}{3pt}
\begin{tabular}{lccccccc@{\hspace{6pt}}cccc@{\hspace{6pt}}c}
\toprule
& \multicolumn{7}{c}{\textbf{Off-the-shelf}} & \multicolumn{4}{c}{\textbf{Retrained \mbox{XPV}}} & \\
\cmidrule(lr){2-8} \cmidrule(lr){9-12}
\textbf{Generator}
& DeFake
& PatchCraft~
& RINE
& SPAI
& B-Free
& FakeVLM
& SIDA
& SPAI
& RINE
& B-Free
& DINO-MAC
& \ours \\
\midrule
\multicolumn{13}{@{}l}{\textit{Stable Diffusion}} \\
SD1.3 & 88.6 & 95.7 & 91.4 & 99.0 & \textbf{100.0} & 69.5 & 80.7 & 80.6 & 82.0 & 97.1 & \underline{99.9} & \textbf{100.0} \\
SD1.4 & 88.9 & 95.7 & 91.5 & 99.0 & \textbf{100.0} & 70.9 & 81.2 & 81.2 & 82.0 & 97.2 & \underline{99.9} & \textbf{100.0} \\
SD2 & 76.3 & 95.5 & 83.6 & 93.3 & \textbf{100.0} & 64.8 & 61.2 & 77.7 & 83.8 & 90.8 & 99.1 & \underline{99.3} \\
SDXL & 84.4 & 96.3 & 86.5 & 94.0 & \textbf{100.0} & 58.4 & 87.5 & 92.0 & 98.5 & 92.3 & \underline{99.9} & \textbf{100.0} \\
SD3 & 95.5 & 32.5 & 30.7 & 61.2 & 92.7 & 33.1 & \textbf{96.6} & 73.2 & 79.5 & 94.5 & 94.0 & \underline{96.1} \\
\addlinespace
\multicolumn{13}{@{}l}{\textit{Midjourney }} \\
Midjourney v5 & 83.5 & 78.0 & 75.7 & 97.4 & \textbf{99.9} & 57.0 & 94.6 & 85.4 & 95.1 & 99.7 & 99.1 & \underline{99.8} \\
Midjourney v6.1 & 86.4 & 95.7 & 67.7 & 80.4 & 96.2 & 52.5 & 96.1 & 93.8 & 95.2 & 94.3 & \underline{98.3} & \textbf{99.0} \\
Midjourney v7 & 81.9 & 94.8 & 63.7 & 83.7 & 92.7 & 49.5 & \underline{95.5} & 79.0 & 86.3 & 92.6 & 78.9 & \textbf{96.6} \\
\addlinespace
\multicolumn{13}{@{}l}{\textit{DALL-E}} \\
DALLE-2 & 68.1 & 80.0 & 75.6 & 83.2 & 98.8 & 64.8 & 61.1 & 71.9 & 89.5 & 99.5 & \underline{99.7} & \textbf{99.8} \\
DALLE-3 & 94.8 & 28.5 & 72.3 & 81.1 & 99.5 & 37.1 & \underline{99.7} & 90.6 & 98.0 & 99.1 & \textbf{100.0} & \textbf{100.0} \\
\addlinespace
\multicolumn{13}{@{}l}{\textit{GPT}} \\
GPT-Image-2 & \underline{82.6} & 59.4 & 29.4 & 63.7 & 73.0 & 36.6 & 82.3 & 62.1 & 74.8 & 78.9 & 81.1 & \textbf{90.7} \\
GPT-Image-1.5 & 80.0 & 78.7 & 42.6 & 78.0 & 66.8 & 46.9 & \underline{85.5} & 73.0 & 76.1 & 83.2 & 81.3 & \textbf{94.9} \\
\addlinespace
\multicolumn{13}{@{}l}{\textit{Flux}} \\
Flux & 95.5 & 67.2 & 68.1 & 80.0 & 93.9 & 32.7 & 99.3 & 91.9 & 93.8 & 98.7 & \underline{99.8} & \textbf{99.9} \\
Flux2-K9 & \textbf{86.5} & 55.4 & 11.9 & 34.5 & 44.5 & 40.0 & \underline{76.3} & 54.3 & 39.6 & 45.3 & 41.0 & 53.6 \\
\addlinespace
\multicolumn{13}{@{}l}{\textit{Video}} \\
Sora2 & 84.3 & 63.7 & 32.3 & 55.0 & 83.0 & 52.5 & 93.4 & 81.9 & 87.2 & 91.7 & \underline{96.4} & \textbf{99.3} \\
Veo3 & 80.9 & 56.1 & 34.7 & 61.1 & 94.0 & 50.9 & 94.4 & 92.8 & 89.1 & \underline{95.6} & 93.9 & \textbf{99.3} \\
Wan2.5 & 88.4 & 50.9 & 33.0 & 65.1 & 91.7 & 43.3 & 96.5 & \underline{98.3} & 96.3 & 95.6 & 97.4 & \textbf{99.9} \\
\addlinespace
\multicolumn{13}{@{}l}{\textit{OpenFake}} \\
Aurora & 91.6 & 79.7 & 32.1 & 88.7 & 90.1 & 38.0 & 96.8 & 94.2 & \underline{97.3} & 96.0 & \textbf{100.0} & \textbf{100.0} \\
Frames & 91.0 & \textbf{97.7} & 70.5 & 85.6 & 88.2 & 57.9 & \underline{96.6} & 80.1 & 84.2 & 89.2 & 80.8 & 91.8 \\
Halfmoon & 87.7 & 88.2 & 77.3 & 95.3 & 94.3 & 55.5 & \underline{95.9} & 78.3 & 79.7 & \textbf{96.1} & 78.8 & 92.9 \\
Lumina & 97.2 & 85.5 & 85.9 & 76.6 & \underline{99.6} & 63.9 & 97.9 & 93.9 & 91.9 & 97.4 & \textbf{99.7} & \underline{99.6} \\
Ideo2 & 89.6 & \underline{97.0} & 66.0 & 91.6 & \textbf{99.4} & 53.8 & 96.9 & 58.3 & 91.3 & 95.9 & 84.3 & 95.8 \\
Recraft2 & 96.1 & 29.4 & 49.9 & 67.8 & \textbf{99.8} & 38.4 & \underline{97.7} & 49.3 & 71.7 & 96.5 & 93.8 & 97.2 \\
Recraft3 & 89.4 & 39.3 & 49.0 & 88.7 & 83.5 & 37.0 & \underline{91.3} & 70.6 & 73.5 & 86.9 & 87.4 & \textbf{95.7} \\
\addlinespace
\multicolumn{13}{@{}l}{\textit{Other}} \\
ERNIE & 87.8 & 64.8 & 43.0 & 72.7 & 94.0 & 43.4 & 93.5 & 95.5 & 93.2 & \underline{96.6} & 91.0 & \textbf{96.7} \\
ERNIE-T & 90.5 & 54.4 & 31.4 & 68.1 & 88.7 & 34.3 & \textbf{96.1} & 90.3 & 90.5 & \underline{94.5} & 84.6 & 93.0 \\
Firefly & 84.5 & 78.2 & 70.7 & 87.6 & \textbf{99.9} & 40.6 & \underline{90.6} & 84.7 & 86.0 & 82.1 & 86.4 & 90.0 \\
GigaGAN & 58.9 & 86.9 & 93.4 & 92.9 & \underline{96.4} & 83.1 & 66.4 & 89.9 & 80.1 & 80.1 & \textbf{99.8} & \textbf{99.8} \\
GLIDE & 89.2 & 76.9 & 83.6 & 86.5 & 95.7 & 88.5 & 74.1 & 97.0 & 73.0 & 92.9 & \textbf{99.9} & \underline{99.8} \\
Illustrious XL & 83.2 & 67.0 & 42.8 & 79.8 & 99.5 & 42.2 & 98.0 & 94.0 & 98.1 & 97.1 & \underline{99.7} & \textbf{99.9} \\
Nano Banana Pro & 71.0 & 59.7 & 33.5 & 60.8 & \underline{96.7} & 56.4 & 91.1 & 94.9 & 93.8 & 92.9 & 94.2 & \textbf{97.3} \\
Seedream & 67.0 & 53.3 & 62.4 & 82.1 & \underline{99.2} & 29.4 & 98.9 & 95.1 & 96.8 & 97.2 & 96.8 & \textbf{99.3} \\
Z-Image & 88.3 & 71.8 & 36.5 & 71.3 & \underline{89.6} & 47.1 & 85.9 & 65.3 & 76.9 & 85.1 & 83.1 & \textbf{95.0} \\
\midrule
\textbf{Mean} & 85.1 & 71.3 & 58.1 & 79.0 & \underline{92.2} & 50.6 & 89.4 & 82.2 & 85.6 & 91.6 & 91.5 & \textbf{96.1} \\
\bottomrule
\end{tabular}
\caption{\textbf{Detection performance on the full generator suite.} AUC ($\times100$) on all 33 generators. Best per row in \textbf{bold}, second best \underline{underlined}, with ties marked jointly. Off-the-shelf baselines are DeFake~\cite{sha2023fake}, PatchCraft~\cite{zhong2024patchcraft}, RINE~\cite{koutlis2024leveraging}, SPAI~\cite{karageorgiou2025spai}, B-Free~\cite{guillaro2024biasfree}, FakeVLM~\cite{wen2025fakevlm} and SIDA~\cite{huang2024sida}; the \mbox{XPV} columns retrain SPAI, RINE, B-Free and DINO-MAC on our training split. The OpenFake group collects generators sourced from the OpenFake benchmark~\cite{livernoche2025openfake}; none of them, and none of the other generators here, appear in any training split used in this work.}
\label{tab:detection-auc}
\end{table}

\begin{table}[t]
\centering
\small
\setlength{\tabcolsep}{3pt}
\begin{tabular}{lccccccc@{\hspace{6pt}}cccc@{\hspace{6pt}}c}
\toprule
& \multicolumn{7}{c}{\textbf{Off-the-shelf}} & \multicolumn{4}{c}{\textbf{Retrained \mbox{XPV}}} & \\
\cmidrule(lr){2-8} \cmidrule(lr){9-12}
\textbf{Generator}
& DeFake
& PatchCraft
& RINE
& SPAI
& B-Free
& FakeVLM
& SIDA
& SPAI
& RINE
& B-Free
& DINO-MAC
& \ours \\
\midrule
\multicolumn{13}{@{}l}{\textit{Stable Diffusion}} \\
SD1.3 & 79.8 & 85.6 & 81.6 & 84.8 & \underline{98.9} & 50.4 & 71.7 & 60.6 & 70.7 & 94.9 & 96.9 & \textbf{99.8} \\
SD1.4 & 80.5 & 85.2 & 81.7 & 84.8 & \underline{98.9} & 50.4 & 72.6 & 61.0 & 71.2 & 95.1 & 96.3 & \textbf{99.7} \\
SD2 & 68.7 & 85.4 & 76.1 & 81.6 & \textbf{98.9} & 50.2 & 57.5 & 59.9 & 72.9 & \underline{90.0} & 77.8 & 87.2 \\
SDXL & 76.8 & 87.9 & 78.8 & 83.3 & \underline{99.0} & 50.2 & 79.2 & 81.3 & 86.4 & 95.7 & 97.5 & \textbf{99.5} \\
SD3 & \underline{85.7} & 38.5 & 38.7 & 53.6 & 78.2 & 47.3 & \textbf{89.1} & 69.7 & 68.5 & 83.3 & 76.5 & 84.1 \\
\addlinespace
\multicolumn{13}{@{}l}{\textit{Midjourney }} \\
Midjourney v5 & 75.2 & 70.6 & 68.6 & 84.6 & \textbf{98.5} & 50.1 & 86.6 & 77.8 & 84.7 & \underline{95.9} & 85.9 & 95.3 \\
Midjourney v6.1 & 77.0 & 84.0 & 63.5 & 71.5 & 88.5 & 38.7 & 87.0 & 87.1 & 82.9 & \textbf{92.6} & 84.1 & \underline{91.0} \\
Midjourney v7 & 74.4 & 85.8 & 58.5 & 74.9 & 77.0 & 49.9 & \underline{87.5} & 72.7 & 77.5 & \textbf{89.5} & 57.3 & 79.2 \\
\addlinespace
\multicolumn{13}{@{}l}{\textit{DALL-E}} \\
DALLE-2 & 61.4 & 71.1 & 68.6 & 75.9 & \underline{93.6} & 50.2 & 57.7 & 52.4 & 81.1 & 68.9 & 89.8 & \textbf{95.7} \\
DALLE-3 & 85.1 & 38.8 & 64.7 & 73.3 & 96.9 & 47.7 & 92.0 & 83.8 & 86.5 & 95.7 & \underline{98.4} & \textbf{99.6} \\
\addlinespace
\multicolumn{13}{@{}l}{\textit{GPT}} \\
GPT-Image-2 & 75.3 & 63.5 & 50.4 & 62.9 & 72.1 & 48.6 & 74.3 & 71.9 & 68.3 & \textbf{76.4} & 68.3 & \underline{76.3} \\
GPT-Image-1.5 & 72.5 & 71.8 & 45.0 & 71.0 & 60.2 & 31.8 & \textbf{77.5} & 69.3 & 64.4 & 63.2 & 61.1 & \underline{72.9} \\
\addlinespace
\multicolumn{13}{@{}l}{\textit{Flux}} \\
Flux & 85.8 & 65.3 & 62.4 & 71.1 & 78.8 & 47.4 & 91.5 & 85.3 & 82.6 & 92.5 & \underline{96.3} & \textbf{98.0} \\
Flux2-K9 & \textbf{78.4} & 48.5 & 35.1 & 38.5 & 51.2 & 49.0 & \underline{68.8} & 59.1 & 41.4 & 49.5 & 50.8 & 53.8 \\
\addlinespace
\multicolumn{13}{@{}l}{\textit{Video}} \\
Sora2 & 75.5 & 59.8 & 48.3 & 56.5 & 78.3 & 35.5 & 85.9 & 79.7 & 76.4 & 83.7 & \underline{87.0} & \textbf{95.6} \\
Veo3 & 72.9 & 49.2 & 40.6 & 53.1 & 81.3 & 49.6 & 87.1 & 83.6 & 79.9 & \underline{87.9} & 73.2 & \textbf{94.3} \\
Wan2.5 & 80.4 & 44.5 & 39.8 & 56.4 & 85.2 & 49.4 & 88.9 & \underline{92.8} & 85.6 & 91.9 & 86.6 & \textbf{98.6} \\
\addlinespace
\multicolumn{13}{@{}l}{\textit{OpenFake}} \\
Aurora & 78.2 & 74.2 & 55.6 & 74.8 & 88.8 & 21.3 & 86.4 & 88.5 & 80.1 & 93.9 & \underline{99.3} & \textbf{99.5} \\
Frames & 77.5 & 80.2 & 67.3 & 72.7 & 85.9 & 20.6 & \underline{86.3} & 82.8 & 75.5 & 84.3 & 82.6 & \textbf{88.8} \\
Halfmoon & 76.1 & 77.2 & 69.8 & 74.3 & \underline{92.4} & 16.6 & 85.3 & 84.5 & 73.1 & \textbf{93.6} & 87.5 & 90.8 \\
Lumina & 83.0 & 77.0 & 75.3 & 69.2 & \underline{97.3} & 35.4 & 88.1 & 88.7 & 79.9 & 95.7 & 94.7 & \textbf{97.8} \\
Ideo2 & 77.4 & 80.8 & 65.4 & 74.8 & \textbf{96.9} & 22.2 & 86.4 & 75.1 & 77.6 & \underline{93.4} & 82.7 & 89.4 \\
Recraft2 & 79.8 & 59.8 & 60.2 & 66.3 & \textbf{98.0} & 21.7 & 86.3 & 73.4 & 67.1 & \underline{93.8} & 85.5 & 91.9 \\
Recraft3 & 81.1 & 39.5 & 46.4 & 80.7 & 72.3 & 49.0 & \textbf{83.5} & 57.7 & 62.8 & 74.2 & 61.7 & \underline{81.8} \\
\addlinespace
\multicolumn{13}{@{}l}{\textit{Other}} \\
ERNIE & 77.0 & 67.5 & 57.6 & 68.5 & 91.6 & 23.9 & 84.7 & 89.8 & 78.9 & \textbf{93.9} & 85.8 & \underline{93.3} \\
ERNIE-T & 80.2 & 54.2 & 44.9 & 64.5 & 82.4 & 39.2 & \underline{88.0} & 86.4 & 78.1 & \textbf{90.7} & 76.5 & 87.7 \\
Firefly & 76.4 & 71.6 & 63.1 & 80.0 & \textbf{98.6} & 48.8 & 82.6 & 71.8 & 80.1 & \underline{90.7} & 56.7 & 58.7 \\
GigaGAN & 56.8 & 78.0 & 82.5 & 83.1 & \underline{87.5} & 50.4 & 64.4 & 71.7 & 71.2 & 68.8 & 87.0 & \textbf{95.2} \\
GLIDE & 80.7 & 69.1 & 76.5 & 80.3 & 84.7 & 50.4 & 65.9 & 92.0 & 62.9 & 83.7 & \underline{95.6} & \textbf{97.5} \\
Illustrious XL & 75.3 & 57.6 & 43.4 & 72.2 & \underline{97.6} & 49.8 & 90.3 & 88.2 & 86.7 & 95.1 & 89.7 & \textbf{98.5} \\
Nano Banana Pro & 69.1 & 59.9 & 53.4 & 60.3 & \underline{92.9} & 28.2 & 83.1 & 88.9 & 80.2 & \textbf{95.2} & 85.0 & 89.3 \\
Seedream & 69.5 & 57.6 & 62.0 & 72.9 & \underline{96.9} & 26.9 & 88.3 & 90.2 & 81.0 & 95.1 & 86.3 & \textbf{97.2} \\
Z-Image & \underline{80.1} & 63.9 & 41.1 & 63.1 & 75.6 & 49.6 & 78.1 & 64.1 & 66.2 & 70.8 & 62.1 & \textbf{83.4} \\
\midrule
\textbf{Mean} & 76.5 & 66.8 & 59.6 & 70.8 & \underline{87.1} & 40.9 & 81.3 & 77.0 & 75.2 & 86.7 & 81.9 & \textbf{89.7} \\
\bottomrule
\end{tabular}
\caption{\textbf{Accuracy at a fixed threshold of 0.5 on the full generator suite.} Accuracy ($\times100$) on all 33 generators. Best per row in \textbf{bold}, second best \underline{underlined}, with ties bolded jointly. Off-the-shelf baselines are DeFake~\cite{sha2023fake}, PatchCraft~\cite{zhong2024patchcraft}, RINE~\cite{koutlis2024leveraging}, SPAI~\cite{karageorgiou2025spai}, B-Free~\cite{guillaro2024biasfree}, FakeVLM~\cite{wen2025fakevlm} and SIDA~\cite{huang2024sida}; the \mbox{XPV} columns retrain SPAI, RINE, B-Free and DINO-MAC on our training split. The OpenFake group collects generators sourced from the OpenFake benchmark~\cite{livernoche2025openfake}; none of them, and none of the other generators here, appear in any training split used in this work.}
\label{tab:detection-acc}
\end{table}

\begin{table}[t]
\centering
\small
\setlength{\tabcolsep}{3pt}
\begin{tabular}{lccccccc@{\hspace{6pt}}cccc@{\hspace{6pt}}c}
\toprule
& \multicolumn{7}{c}{\textbf{Off-the-shelf}} & \multicolumn{4}{c}{\textbf{Retrained \mbox{XPV}}} & \\
\cmidrule(lr){2-8} \cmidrule(lr){9-12}
\textbf{Generator}
& DeFake
& PatchCraft
& RINE
& SPAI
& B-Free
& FakeVLM
& SIDA
& SPAI
& RINE
& B-Free
& DINO-MAC
& \textbf{\ours} \\
\midrule
\multicolumn{13}{@{}l}{\textit{Stable Diffusion}} \\
SD1.3 & 88.0 & 96.2 & 90.3 & 98.1 & \textbf{100.0} & 70.3 & 82.0 & 77.2 & 85.1 & 97.2 & \underline{99.9} & \textbf{100.0} \\
SD1.4 & 88.3 & 96.3 & 90.1 & 98.1 & \textbf{100.0} & 71.1 & 82.4 & 77.9 & 85.0 & 97.2 & \underline{99.9} & \textbf{100.0} \\
SD2 & 76.0 & 95.9 & 80.7 & 93.3 & \textbf{100.0} & 68.6 & 64.5 & 76.0 & 86.7 & 95.9 & 99.1 & \underline{99.2} \\
SDXL & 84.6 & 94.7 & 84.0 & 92.9 & \textbf{100.0} & 60.2 & 89.6 & 92.3 & 98.8 & 97.3 & \underline{99.9} & \textbf{100.0} \\
SD3 & 95.1 & 41.1 & 39.0 & 58.5 & 93.7 & 39.5 & \textbf{97.2} & 79.4 & 81.3 & 93.4 & 95.4 & \underline{97.0} \\
\addlinespace
\multicolumn{13}{@{}l}{\textit{Midjourney }} \\
Midjourney v5 & 82.7 & 79.8 & 71.0 & 96.5 & \textbf{99.9} & 58.3 & 95.6 & 87.8 & 95.6 & 97.4 & 99.2 & \underline{99.8} \\
Midjourney v6.1 & 81.8 & 93.5 & 56.4 & 75.1 & 95.3 & 43.9 & 95.5 & 92.6 & 93.9 & 95.8 & \underline{97.9} & \textbf{98.6} \\
Midjourney v7 & 82.6 & 94.5 & 62.3 & 83.2 & 93.5 & 50.3 & 95.5 & 82.7 & 86.2 & \underline{95.9} & 86.3 & \textbf{97.2} \\
\addlinespace
\multicolumn{13}{@{}l}{\textit{DALL-E}} \\
DALLE-2 & 67.2 & 78.0 & 72.0 & 81.6 & 99.0 & 64.4 & 64.4 & 66.9 & 89.3 & 89.0 & \underline{99.7} & \textbf{99.8} \\
DALLE-3 & 94.6 & 39.4 & 67.2 & 78.2 & 99.6 & 43.5 & \underline{99.7} & 92.2 & 98.3 & 97.3 & \textbf{100.0} & \textbf{100.0} \\
\addlinespace
\multicolumn{13}{@{}l}{\textit{GPT}} \\
GPT-Image-2 & 70.1 & 50.3 & 23.8 & 47.3 & 62.9 & 42.3 & \underline{82.3} & 56.4 & 64.0 & 74.7 & 78.4 & \textbf{85.3} \\
GPT-Image-1.5 & 78.9 & 80.7 & 45.9 & 78.8 & 72.5 & 35.1 & 85.5 & 79.0 & 77.8 & 79.8 & \underline{89.0} & \textbf{95.8} \\
\addlinespace
\multicolumn{13}{@{}l}{\textit{Flux}} \\
Flux & 95.2 & 73.2 & 67.5 & 82.4 & 94.5 & 40.3 & 99.5 & 93.7 & 95.0 & 96.7 & \underline{99.8} & \textbf{99.9} \\
Flux2-K9 & \textbf{85.8} & 55.8 & 32.5 & 41.6 & 50.6 & 44.1 & \underline{79.2} & 64.3 & 46.4 & 52.2 & 49.6 & 61.3 \\
\addlinespace
\multicolumn{13}{@{}l}{\textit{Video}} \\
Sora2 & 75.5 & 51.5 & 27.5 & 41.8 & 80.5 & 41.6 & 92.6 & 78.0 & 81.8 & 89.2 & \underline{96.2} & \textbf{98.6} \\
Veo3 & 81.3 & 56.3 & 41.1 & 57.9 & 95.0 & 54.5 & 95.3 & 93.0 & 89.9 & \underline{95.5} & 95.3 & \textbf{99.3} \\
Wan2.5 & 88.2 & 52.3 & 39.9 & 60.9 & 94.2 & 47.3 & 97.3 & \underline{98.2} & 96.8 & 96.0 & 98.1 & \textbf{99.9} \\
\addlinespace
\multicolumn{13}{@{}l}{\textit{OpenFake}} \\
Aurora & 79.5 & 52.5 & 16.0 & 65.8 & 81.5 & 17.2 & 93.7 & 86.3 & 92.1 & \underline{94.3} & \textbf{99.9} & \textbf{99.9} \\
Frames & 75.4 & \underline{92.1} & 37.1 & 61.0 & 73.4 & 29.0 & \textbf{92.4} & 65.9 & 57.0 & 74.3 & 65.7 & 80.5 \\
Halfmoon & 64.6 & 68.1 & 38.1 & 78.0 & 84.7 & 20.9 & \underline{88.8} & 62.1 & 51.4 & \textbf{91.9} & 64.1 & 80.0 \\
Lumina & 95.1 & 80.0 & 75.1 & 66.5 & \underline{99.5} & 52.3 & 97.1 & 93.1 & 87.9 & 97.3 & \textbf{99.6} & 99.3 \\
Ideo2 & 73.5 & 91.0 & 33.5 & 74.6 & \textbf{98.2} & 24.9 & 92.8 & 37.6 & 80.6 & \underline{94.2} & 74.1 & 90.9 \\
Recraft2 & 87.9 & 16.7 & 23.8 & 40.3 & \textbf{99.3} & 18.0 & \underline{95.3} & 31.1 & 44.7 & 94.5 & 88.3 & 91.9 \\
Recraft3 & 89.8 & 44.7 & 49.4 & 86.8 & 87.1 & 41.5 & \underline{92.9} & 69.2 & 71.9 & 88.8 & 89.5 & \textbf{95.9} \\
\addlinespace
\multicolumn{13}{@{}l}{\textit{Other}} \\
ERNIE & 72.8 & 48.1 & 22.0 & 53.0 & 89.5 & 24.8 & 89.1 & 92.5 & 86.3 & \textbf{95.1} & 87.5 & \underline{93.5} \\
ERNIE-T & 87.5 & 51.7 & 31.1 & 61.1 & 88.9 & 31.5 & \textbf{96.1} & 91.4 & 90.2 & \underline{94.5} & 87.0 & 93.3 \\
Firefly & 83.4 & 76.9 & 66.2 & 85.4 & \textbf{99.9} & 44.6 & 92.2 & 83.8 & 82.6 & \underline{95.7} & 92.5 & 90.4 \\
GigaGAN & 61.0 & 88.6 & 92.9 & 91.6 & 97.0 & 83.0 & 71.6 & 85.4 & 85.1 & 84.9 & \underline{99.7} & \textbf{99.8} \\
GLIDE & 89.2 & 78.1 & 79.8 & 82.8 & 96.3 & 88.7 & 74.6 & 95.9 & 76.9 & 93.6 & \textbf{99.9} & \underline{99.8} \\
Illustrious XL & 82.4 & 69.3 & 45.5 & 80.5 & 99.6 & 43.8 & 98.3 & 95.5 & 98.3 & 97.2 & \underline{99.7} & \textbf{99.8} \\
Nano Banana Pro & 51.5 & 39.8 & 21.3 & 37.6 & 94.8 & 40.8 & 91.1 & 89.9 & 87.3 & \textbf{96.3} & 91.7 & \underline{95.4} \\
Seedream & 54.9 & 33.3 & 35.3 & 61.9 & \textbf{98.5} & 18.9 & 97.6 & 94.4 & 91.8 & 96.5 & 95.3 & \underline{98.2} \\
Z-Image & 87.6 & 71.1 & 42.6 & 69.8 & \underline{91.5} & 49.9 & 88.7 & 72.9 & 77.4 & 87.1 & 87.2 & \textbf{95.7} \\
\midrule
\textbf{Mean} & 80.4 & 67.6 & 51.5 & 71.6 & 91.2 & 45.6 & 89.4 & 79.8 & 82.2 & \underline{91.4} & 91.1 & \textbf{95.0} \\
\bottomrule
\end{tabular}
\caption{\textbf{Average precision on the full generator suite.} AP ($\times100$) on all 33 generators. Best per row in \textbf{bold}, second best \underline{underlined}, with ties bolded jointly. Off-the-shelf baselines are DeFake~\cite{sha2023fake}, PatchCraft~\cite{zhong2024patchcraft}, RINE~\cite{koutlis2024leveraging}, SPAI~\cite{karageorgiou2025spai}, B-Free~\cite{guillaro2024biasfree}, FakeVLM~\cite{wen2025fakevlm} and SIDA~\cite{huang2024sida}; the \mbox{XPV} columns retrain SPAI, RINE, B-Free and DINO-MAC on our training split. The OpenFake group collects generators sourced from the OpenFake benchmark~\cite{livernoche2025openfake}; none of them, and none of the other generators here, appear in any training split used in this work.}
\label{tab:detection-ap}
\end{table}

\begin{figure*}[p]
\centering
\small
\setlength{\tabcolsep}{3pt}
\renewcommand{\arraystretch}{1.05}
\begin{tabular}{@{}c >{\centering\arraybackslash}m{0.15\linewidth} m{0.29\linewidth} >{\centering\arraybackslash}m{0.15\linewidth} m{0.29\linewidth}@{}}
  \noalign{\vskip 18pt}
  \rotatebox[origin=c]{90}{\footnotesize\textbf{GPT-Image-1.5}}
  & \includegraphics[width=\linewidth,height=2.9cm,keepaspectratio]{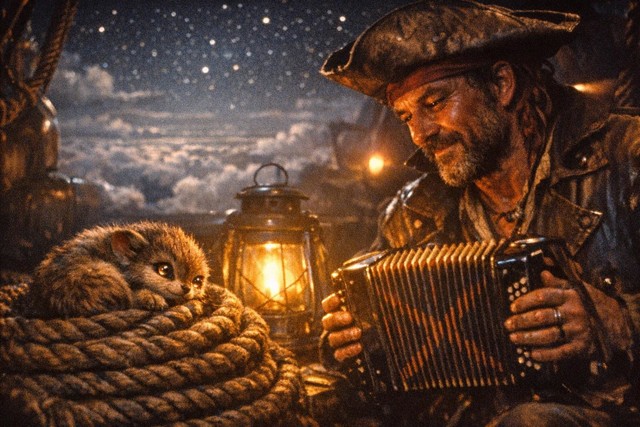}
  & \scriptsize\textbf{\ours: Verdict: fake (97.3\%)} $\cdot$ \textbf{Explanation:} Visual inconsistencies appear around the man's face and the accordion. His facial features are unnaturally smoothed and lack the fine skin texture seen on his hands, giving a `plastic' look. Blurring is visible where his jawline meets the background, and the lighting on his face does not match the warm glow from the lantern behind him.
  & \includegraphics[width=\linewidth,height=2.9cm,keepaspectratio]{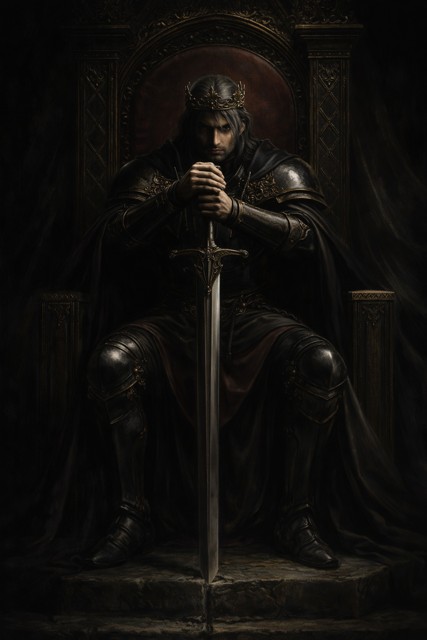}
  & \scriptsize\textbf{\ours: Verdict: fake (95.9\%)} $\cdot$ \textbf{Explanation:} The anatomy of the hands gripping the sword is the most prominent issue: the fingers appear fused and lack distinct knuckles, blending into a single smooth mass rather than showing individual digits. The sword itself has an overly uniform texture, closer to a digital render than a metallic object. \\[3pt]
  \noalign{\vskip 18pt}
  \rotatebox[origin=c]{90}{\footnotesize\textbf{GPT-Image-2}}
  & \includegraphics[width=\linewidth,height=2.9cm,keepaspectratio]{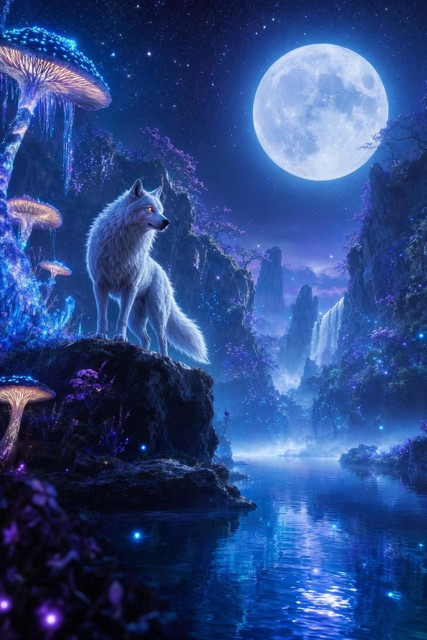}
  & \scriptsize\textbf{\ours: Verdict: fake (94.4\%)} $\cdot$ \textbf{Explanation:} The white wolf has a soft, airbrushed texture that lacks the fine fur detail present in the surrounding environment. The lighting on its body is inconsistent with the deep blue ambient light of the night sky, and it casts no realistic shadow onto the rock beneath it.
  & \includegraphics[width=\linewidth,height=2.9cm,keepaspectratio]{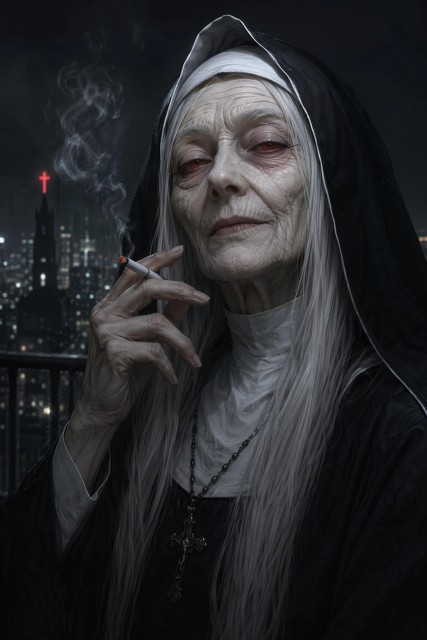}
  & \scriptsize\textbf{\ours: Verdict: fake (67.7\%)} $\cdot$ \textbf{Explanation:} Anatomical and physical inconsistencies suggest manipulation. The woman's hand holding the cigarette is distorted, with fingers that lack natural joints. The cigarette is held at an impossible angle and appears to float or clip through the fingers rather than being firmly grasped. \\[3pt]
  \noalign{\vskip 18pt}
  \rotatebox[origin=c]{90}{\footnotesize\textbf{Midjourney-7}}
  & \includegraphics[width=\linewidth,height=2.9cm,keepaspectratio]{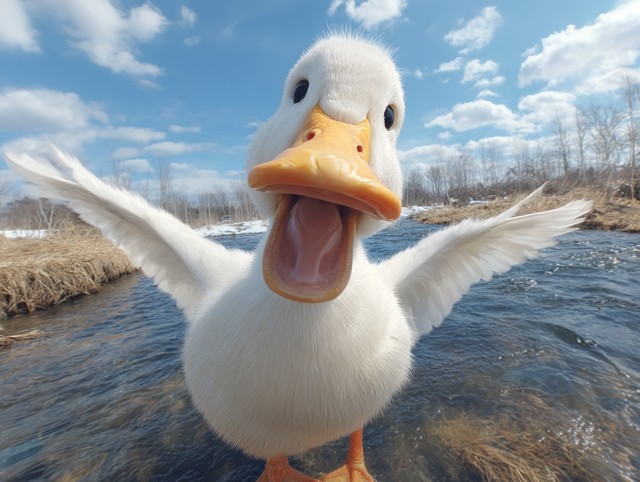}
  & \scriptsize\textbf{\ours: Verdict: fake (99.97\%)} $\cdot$ \textbf{Explanation:} Anatomical and blending inconsistencies are present throughout. The duck's head is disproportionately large for its body, and the neck lacks natural muscle definition where it meets the torso. A halo effect and soft blurring surround the edges of the wings.
  & \includegraphics[width=\linewidth,height=2.9cm,keepaspectratio]{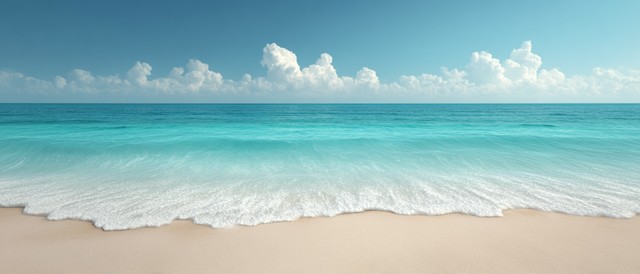}
  & \scriptsize\textbf{\ours: Verdict: fake (99.95\%)} $\cdot$ \textbf{Explanation:} The sky and horizon line carry the clearest synthetic cues. The clouds are flat, repetitive white shapes with no natural texture or depth, creating a `pasted-on' effect. The water's surface lacks realistic wave patterns and reads as a solid, painted layer. \\[3pt]
  \noalign{\vskip 18pt}
  \rotatebox[origin=c]{90}{\footnotesize\textbf{Veo-3}}
  & \includegraphics[width=\linewidth,height=2.9cm,keepaspectratio]{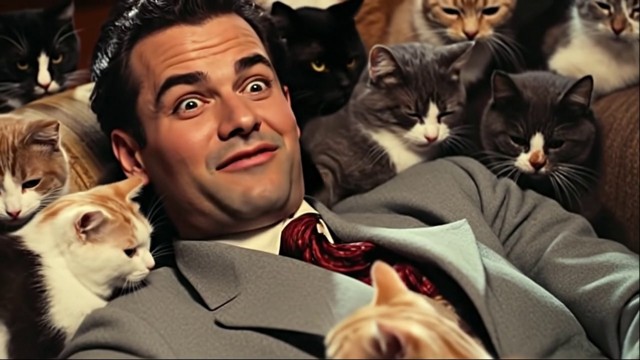}
  & \scriptsize\textbf{\ours: Verdict: fake (99.96\%)} $\cdot$ \textbf{Explanation:} The man's eyes are unnaturally wide and bulging, with a noticeable loss of skin texture around the eyelids. The cats' fur overlapping his jacket lacks realistic depth and appears as flat overlays rather than strands sitting on fabric.
  & \includegraphics[width=\linewidth,height=2.9cm,keepaspectratio]{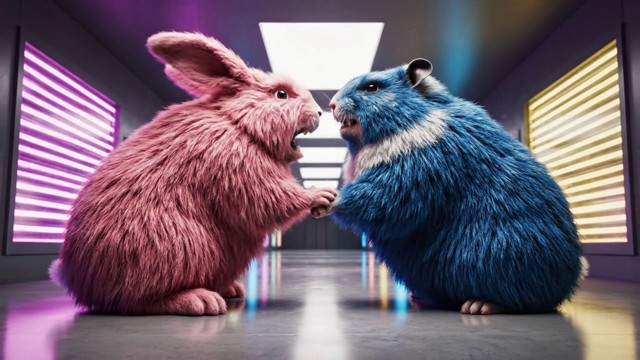}
  & \scriptsize\textbf{\ours: Verdict: fake (99.2\%)} $\cdot$ \textbf{Explanation:} Lighting and texture on both subjects show synthetic qualities. The fur has an unnaturally uniform, brush-like appearance, and the floor reflection lacks the complex distortion expected from such large, textured bodies. \\[3pt]
  \noalign{\vskip 18pt}
  \rotatebox[origin=c]{90}{\footnotesize\textbf{Nano Banana Pro}}
  & \includegraphics[width=\linewidth,height=2.9cm,keepaspectratio]{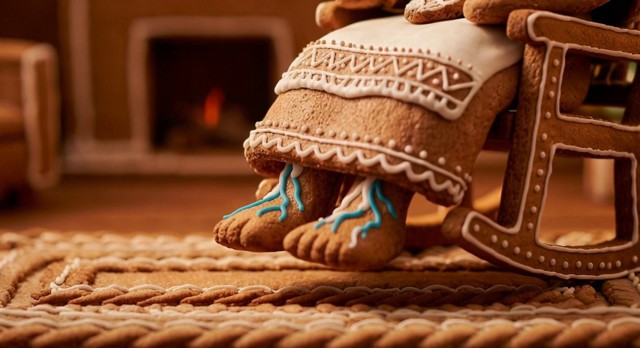}
  & \scriptsize\textbf{\ours: Verdict: fake (80.4\%)} $\cdot$ \textbf{Explanation:} Visual inconsistencies cluster around the gingerbread sled. The white decorative trim is unnaturally flat, and the blue icing on the boots has a digital, painted-on quality that does not follow the natural contours of the dough.
  & \includegraphics[width=\linewidth,height=2.9cm,keepaspectratio]{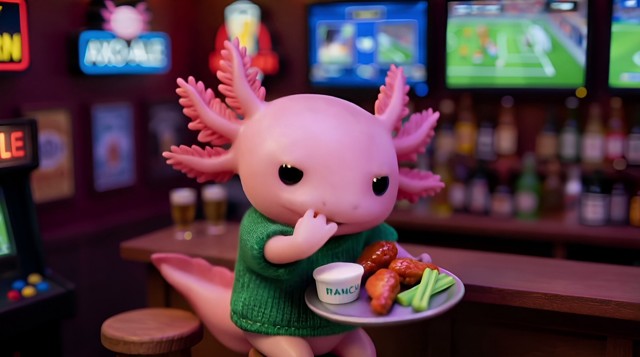}
  & \scriptsize\textbf{\ours: Verdict: fake (99.3\%)} $\cdot$ \textbf{Explanation:} The plate of food floats without realistic contact shadows against the surface below it. The text on the white cup is garbled and illegible, a common artifact in synthetic imagery. \\[3pt]
  \noalign{\vskip 18pt}
\end{tabular}
\caption{\textbf{\ours\ generalizes to unseen, recent generators without retraining.} Each row pairs two examples from one generator with the verdict and explanation produced by \ours. None of the five generators were seen during training, and all verdicts shown are correct.}
\label{fig:appendix_openfake}
\end{figure*}
\begin{figure*}[t]
\centering
\small
\setlength{\tabcolsep}{4pt}
\renewcommand{\arraystretch}{1.1}
\begin{tabular}{@{}c >{\centering\arraybackslash}m{0.16\linewidth} m{0.36\linewidth} m{0.36\linewidth}@{}}
  & \multicolumn{1}{c}{\footnotesize Image}
  & \multicolumn{1}{c}{\footnotesize \ours\ (ours)}
  & \multicolumn{1}{c}{\footnotesize Ground Truth} \\[4pt]
  \rotatebox[origin=c]{90}{\footnotesize\textbf{Fake}}
  & \includegraphics[width=\linewidth,height=3.0cm,keepaspectratio]{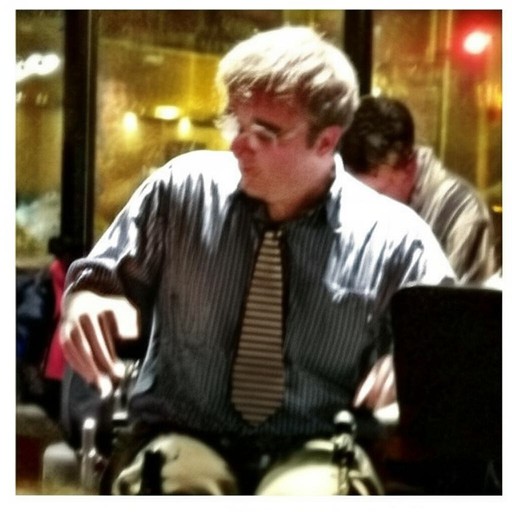}
  & \scriptsize The image exhibits several signs of synthetic generation. The man's right hand, resting on his lap, appears distorted with fingers that seem to merge into a single, indistinct mass rather than showing clear individual digits. The texture of the skin on his face and hands has an overly smooth, plastic-like quality, lacking natural pores or fine detail. Additionally, the background elements, particularly the person behind him, appear smeared and lack coherent structure, blending unnaturally into the surroundings. The lighting on the subject feels somewhat flat and inconsistent with the ambient light sources visible in the window behind him.\par\vspace{2\baselineskip}\noindent(BERTScore F1: 0.75 $\cdot$ E5: 0.92 $\cdot$ Generated tokens: 119)
  & \scriptsize The image exhibits several synthetic qualities, most notably in the rendering of the man's hair, which appears as a blurry, undifferentiated mass that lacks natural strand definition and seems to float unnaturally above his forehead. His glasses are distorted, with the frames blending directly into his skin and hair without clear structural edges. Additionally, the man's right hand is anatomically incorrect, featuring elongated, smudged fingers that lack joints and appear to melt into the surrounding space. The lighting on his face is also inconsistent with the rest of the scene, appearing overly bright and flat compared to the high-contrast shadows on his shirt and the background. \\
  \addlinespace[10pt]
  \rotatebox[origin=c]{90}{\footnotesize\textbf{Fake}}
  & \includegraphics[width=\linewidth,height=3.0cm,keepaspectratio]{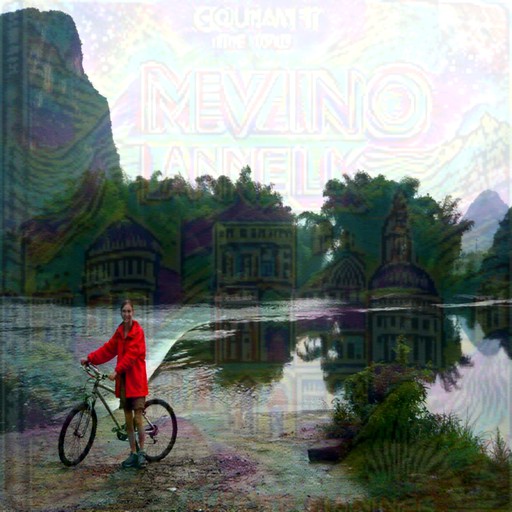}
  & \scriptsize The image contains several visual inconsistencies that suggest manipulation, particularly around the man in the red jacket. His figure appears to be floating slightly above the ground, as there are no contact shadows beneath his boots to ground him in the environment. The bicycle he is holding lacks a realistic connection to his hands; the handlebars seem to merge into his gloves without clear grip points, and the front wheel of the bike has a distorted, smudged appearance that doesn't match the sharpness of the surrounding gravel. Additionally, the lighting on the man's face and jacket is much flatter and more diffused than the high-contrast, directional light hitting the buildings and water in the background, creating a disjointed sense of depth.\par\vspace{2\baselineskip}\noindent(BERTScore F1: 0.65 $\cdot$ E5: 0.90 $\cdot$ Generated tokens: 145)
  & \scriptsize The image contains a significant lighting inconsistency centered on the person's bright red jacket. While the overall scene is overcast with soft, muted lighting and diffused shadows, the red jacket appears unnaturally vibrant and lacks the atmospheric haze present on the rest of the subject and the background. Furthermore, the reflection in the water directly behind the person does not accurately represent the intense red color of the garment, showing a much duller, desaturated tone that contradicts the jacket's current appearance. \\
  \addlinespace[10pt]
  \rotatebox[origin=c]{90}{\footnotesize\textbf{Real}}
  & \includegraphics[width=\linewidth,height=3.0cm,keepaspectratio]{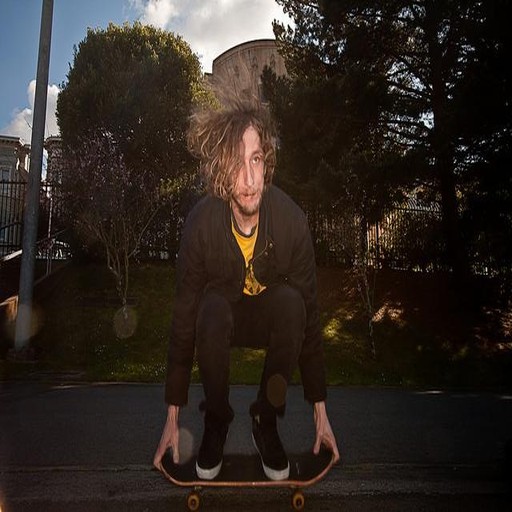}
  & \scriptsize This picture looks real because the boy's hair is all messy and flying up in the air. It shows he was really moving fast on his skateboard, just like when you jump or turn really quickly.\par\vspace{2\baselineskip}\noindent(BERTScore F1: 0.79 $\cdot$ E5: 0.95 $\cdot$ Generated tokens: 41)
  & \scriptsize This picture looks real because the man's hair is all messy and flying up in the air. It looks like he is really moving or jumping on his skateboard, and the wind is making his hair go whoosh! \\
  \addlinespace[10pt]
  \rotatebox[origin=c]{90}{\footnotesize\textbf{Real}}
  & \includegraphics[width=\linewidth,height=3.0cm,keepaspectratio]{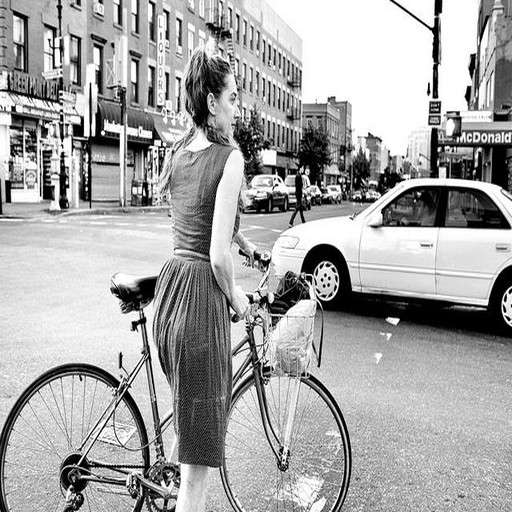}
  & \scriptsize This picture looks real because of the bike's little basket. You can see how it's holding a big, wrinkly plastic bag full of stuff, just like someone would carry on their way to work or run errands.\par\vspace{2\baselineskip}\noindent(BERTScore F1: 0.65 $\cdot$ E5: 0.85 $\cdot$ Generated tokens: 46)
  & \scriptsize This picture looks real because the street isn't perfectly clean. You can see some little pieces of paper on the road by the car's tire, which happens in a real city but is the kind of tiny mess a computer might forget to draw. \\
\end{tabular}
\caption{\textbf{Correct, forensically grounded explanations from \ours\ across the BERTScore distribution on XPV.} The rotated label gives the ground-truth class; \ours\ predicts the correct verdict in all four cases. The two rows per class are drawn from the high and low ends of the BERTScore distribution.}
\label{fig:appendix_pure_success}
\end{figure*}
\begin{figure*}[t]
\centering
\small
\setlength{\tabcolsep}{4pt}
\renewcommand{\arraystretch}{1.1}
\begin{tabular}{@{}c >{\centering\arraybackslash}m{0.16\linewidth} m{0.36\linewidth} m{0.36\linewidth}@{}}
  & \multicolumn{1}{c}{\footnotesize Image}
  & \multicolumn{1}{c}{\footnotesize \ours\ (ours)}
  & \multicolumn{1}{c}{\footnotesize Ground Truth} \\[4pt]
  \rotatebox[origin=c]{90}{\footnotesize\textbf{Fake $\rightarrow$ Real}}
  & \includegraphics[width=\linewidth,height=3.0cm,keepaspectratio]{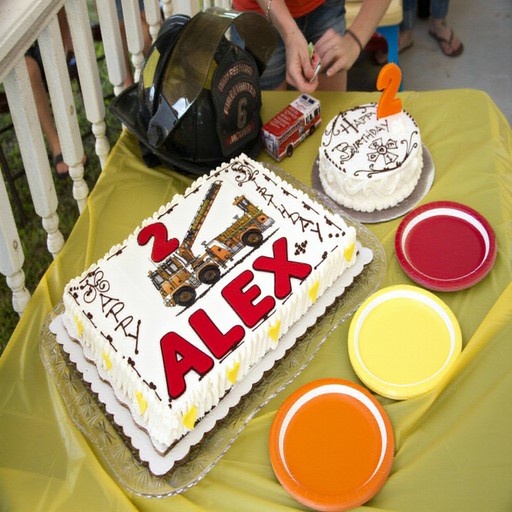}
  & \scriptsize This picture looks real because the writing on the big fire truck cake isn't perfect. You can see how the frosting is a little bumpy and follows the shape of the letters, just like when a real person carefully squeezes it out of a bag to write a name.\par\vspace{2\baselineskip}\noindent(BERTScore F1: 0.60 $\cdot$ E5: 0.88 $\cdot$ Generated tokens: 56)
  & \scriptsize The image contains several visual inconsistencies that suggest digital manipulation. Most notably, the name `ALEX' on the large rectangular cake appears as a flat, two-dimensional graphic rather than physical icing; it lacks the natural highlights, shadows, and texture found on the `2' and the fire truck illustration above it. The red color of the letters is unnaturally uniform and saturated, and the black outlines are too sharp and clean compared to the softer, hand-drawn quality of the surrounding `Happy Birthday' script. Additionally, the letters do not follow the slight perspective or surface irregularities of the cake's frosting, making them look like they are floating on top of the image. \\
  \addlinespace[10pt]
  \rotatebox[origin=c]{90}{\footnotesize\textbf{Fake $\rightarrow$ Real}}
  & \includegraphics[width=\linewidth,height=3.0cm,keepaspectratio]{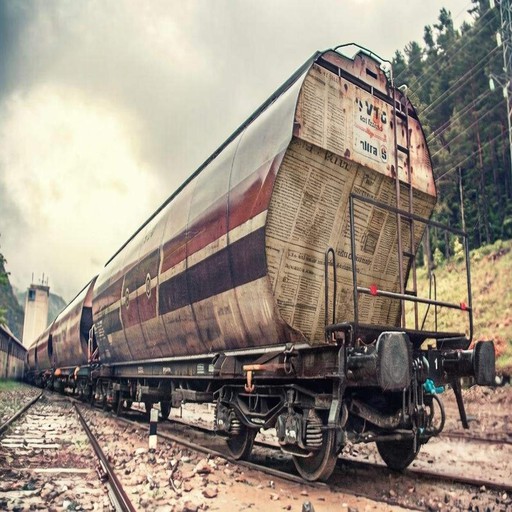}
  & \scriptsize This looks like a real train because its big, round wheels are covered in dirt and mud. You can see how the wheels have been rolling over the rocky ground, which makes them look messy just like a real one would.\par\vspace{2\baselineskip}\noindent(BERTScore F1: 0.51 $\cdot$ E5: 0.84 $\cdot$ Generated tokens: 46)
  & \scriptsize The rear vertical end of the freight car clearly looks synthetic: the curved face is filled with a repeating newspaper column texture whose vertical column edges remain unnaturally straight across the curved surface. The paper pattern does not follow surface wear or rust; text blocks keep uniform contrast and crisp edges while the surrounding metal shows mottled dirt and scratches, creating a visible mismatch in texture and aging. There is a soft halo and slight color fringing where the newspaper meets the metal flange and ladder, indicating a pasted-on layer rather than integrated paint. Several headlines and lines of type are cut off or wrapped oddly at the curved corners, producing impossible letter distortions and partial characters near the left and right edges of the end panel. \\
  \addlinespace[10pt]
  \rotatebox[origin=c]{90}{\footnotesize\textbf{Real $\rightarrow$ Fake}}
  & \includegraphics[width=\linewidth,height=3.0cm,keepaspectratio]{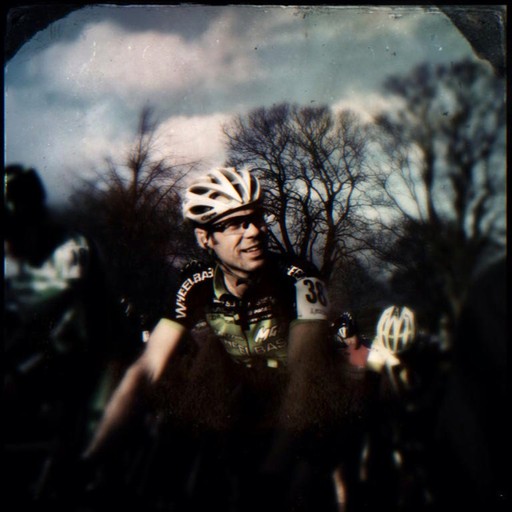}
  & \scriptsize The generated image looks fake due to several visual inconsistencies. Around the cyclist's face, the skin texture appears overly smooth and plastic-like, lacking natural detail. The facial expression itself seems unnatural, an odd, forced smile that doesn't match the rest of the scene's tone. Additionally, the lighting on the face feels inconsistent with the surrounding environment, creating a subtle but noticeable mismatch in brightness and shadow direction. These combined cues make the image appear synthetic rather than authentic.\par\vspace{2\baselineskip}\noindent(BERTScore F1: 0.62 $\cdot$ E5: 0.84 $\cdot$ Generated tokens: 96)
  & \scriptsize This picture looks like a real snapshot because the edges are all dark and fuzzy, like an old photo that got a little dirty. Also, the man's face is a bit blurry because he was moving fast while the picture was taken. \\
  \addlinespace[10pt]
  \rotatebox[origin=c]{90}{\footnotesize\textbf{Real $\rightarrow$ Fake}}
  & \includegraphics[width=\linewidth,height=3.0cm,keepaspectratio]{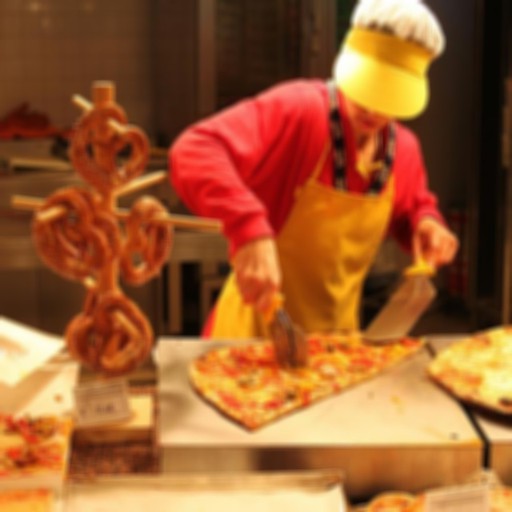}
  & \scriptsize The image contains several visual inconsistencies that suggest manipulation, particularly around the chef's head and hat. The white chef's hat appears unnaturally bright and lacks the soft lighting and shadows found on the rest of the person's body. Furthermore, the yellow band of the hat has a jagged, poorly blended edge where it meets the forehead, and the texture of the hat itself looks overly smooth and flat compared to the more detailed fabric of the red jacket. There is also a slight misalignment in the perspective of the hat's brim relative to the angle of the chef's face.\par\vspace{2\baselineskip}\noindent(BERTScore F1: 0.55 $\cdot$ E5: 0.79 $\cdot$ Generated tokens: 117)
  & \scriptsize This looks like a real picture because you can see how the round pizza cutter is pressing down into the cheesy pizza. It makes a little line and squishes the toppings, just like it would if you were really cutting a warm slice for a snack. \\
\end{tabular}
\caption{\textbf{Representative failure modes of \ours\ on XPV.} The rotated label gives the ground-truth class and the predicted class; the verdict is wrong in all four cases. The two rows per direction are drawn from the high and low ends of the BERTScore distribution among wrong verdict predictions.}
\label{fig:appendix_pure_failure}
\end{figure*}

\end{document}